\documentclass[journal]{IEEEtran}

\usepackage{setspace}
\usepackage{graphicx}
\usepackage{epstopdf}
\usepackage{amsmath, amsthm, amssymb, amsfonts, bm, mathtools, bbm}
\usepackage{amsthm}
\usepackage{amsmath, amssymb, amsthm}
\usepackage{cases, multirow}
\usepackage{accents}
\usepackage{cite}
\usepackage{enumitem}
\usepackage{courier}
\usepackage{subfigure}
\usepackage{float}
\usepackage{color}
\usepackage{adjustbox}
\usepackage{multicol}
\usepackage{dblfloatfix}
\usepackage{cuted, nccmath}
\usepackage{lipsum}
\usepackage{mathtools}
\usepackage{threeparttable}
\usepackage{tabularx}
\usepackage{mdframed}

\usepackage{listings}
\usepackage{color}

\definecolor{dkgreen}{rgb}{0,0.6,0}
\definecolor{gray}{rgb}{0.5,0.5,0.5}
\definecolor{mauve}{rgb}{0.58,0,0.82}

\usepackage{esint}

\long\def\comment#1{}

\def\quad{\hskip0.1em\relax}

\theoremstyle{definition}
\newmdtheoremenv{framed_theo}{Theorem}
\newmdtheoremenv{framed_lem}{Lemma}
\newmdtheoremenv{framed_def}{Definition}
\newmdtheoremenv{framed_cor}{Corollary}
\newmdtheoremenv{framed_remark}{Remark}

\newtheorem{thm}{Theorem}

\newtheorem{corollary}{Corollary}

\def\figref#1{Fig.~\ref{#1}}

\def\be{\begin{equation} }
\def\ee{\end{equation}}

\begin{document}
\title{To Trust or Not to Trust: On Calibration in ML-based Resource Allocation for Wireless Networks}

\author{Rashika Raina,~\IEEEmembership{Student Member,~IEEE}, Nidhi Simmons,~\IEEEmembership{Senior Member,~IEEE}, David E. Simmons, Michel~Daoud~Yacoub,~\IEEEmembership{Member,~IEEE}, and Trung Q. Duong,~\IEEEmembership{Fellow,~IEEE}

\thanks{R. Raina and N. Simmons are with the Centre for Wireless Innovation,  Queen's University of Belfast, Belfast, BT3 9DT, UK. E-mail: \{rraina01, nidhi.simmons\}@qub.ac.uk.}

\thanks{D.E. Simmons is with Dhali Holdings Ltd., Belfast BT5 7HW, UK. E-mail: dr.desimmons@gmail.com.} 

\thanks{M. D. Yacoub is with the School of Electrical and Computer Engineering, University of Campinas, Campinas 13083-970, Brazil. E-mail: mdyacoub@unicamp.br.}

\thanks{T. Q. Duong is with the Faculty of Engineering and Applied Science,
Memorial University, St. John’s, NL A1C 5S7, Canada, and also with the School of
Electronics, Electrical Engineering and Computer Science, Queen’s University
Belfast, Belfast, U.K. E-mail: tduong@mun.ca.}

\thanks{This work was supported by the Royal Academy of Engineering (Grant ref RF\textbackslash201920\textbackslash 19\textbackslash 191). The work of T. Q. Duong was supported in part by the Canada Excellence Research Chair (CERC) Program CERC-2022-00109 and in part by the Natural Sciences and Engineering Research Council of Canada (NSERC) Discovery Grant Program RGPIN-2025-04941.}

\thanks{This paper has been accepted in part for presentation at the IEEE International Conference on Machine Learning for Communication and Networking (ICMLCN), Barcelona, Spain, May 2025~\cite{11140554} and IEEE Global Communications Conference (GLOBECOM) Workshops, Taipei, Taiwan, December 2025~\cite{raina2025ml}. }
}
\maketitle

\begin{abstract} In the next generation communications and networks, machine learning (ML) models are expected to deliver not only highly accurate predictions, but also well-calibrated confidence scores that reflect the true likelihood of correct decisions. In this paper, we study the calibration performance of an ML-based outage predictor within a single-user, multi-resource allocation framework. We begin by establishing key theoretical properties of this system’s outage probability (OP) under perfect calibration. Importantly, we show that as the number of resources grows, the OP of a perfectly calibrated predictor approaches the expected output conditioned on it being below the classification threshold. In contrast, when only a single resource is available, the system's OP equals the model’s overall expected output.
We then derive the OP conditions for a perfectly calibrated predictor. 
These findings guide the choice of the classification threshold to achieve a desired OP, helping system designers meet specific reliability requirements. We further demonstrate that post-processing calibration cannot improve the system's minimum achievable OP, as it does not introduce additional information about future channel states. Additionally, we show that well-calibrated models are part of a broader class of predictors that necessarily improve OP. In particular, we establish a monotonicity condition that the accuracy-confidence function must satisfy for such improvement to occur. To demonstrate these theoretical properties, we conduct a rigorous simulation-based analysis using post-processing calibration techniques, namely, Platt scaling and isotonic regression. As part of this framework, the predictor is trained using an outage loss function specifically designed for this system. Furthermore, this analysis is performed on Rayleigh fading channels with temporal correlation captured by Clarke’s 2D model, which accounts for receiver mobility. Notably, the outage investigated refers to the required resource failing to achieve the transmission capacity requested by the user. 
\end{abstract}

\section{Introduction}
\IEEEPARstart{H}{ow} should we interpret a neural network's (NN) confidence score? If a NN assigns a confidence of 0.4 to the question ``Will this communication channel fail in the future?" what does that really mean? What if it instead returns 0.000001? These questions highlight the importance of \textit{calibration}—the extent to which a network's confidence estimates align with the true probability of an event occurring. In essence, in machine learning (ML)-based systems, calibration concerns the process of adjusting a model's predicted probabilities to correctly reproduce the true likelihood of an event.

A well-calibrated NN allows us to treat its outputs as reliable probabilities, making its predictions more actionable in real-world decision-making. For instance, suppose a user can tolerate a channel failure rate of at most 1 in 100 transmissions. They could consult the NN before using a channel, and if the model predicts a failure probability below 0.01, the channel is deemed acceptable; otherwise, they should seek an alternative. However, this approach is only effective if the NN is well-calibrated, particularly around the critical threshold of 0.01.

Accurate probability estimates are essential for informed decisions, yet NNs often produce confidence scores that do not accurately reflect true probabilities, a phenomenon known as poor calibration. When a model is miscalibrated, it may systematically overestimate or underestimate the likelihood of an event, leading to unreliable decision-making. 
Proper calibration aligns predicted probabilities with actual outcomes, enhancing both the reliability and usability of ML-driven predictions~\cite{guo2017calibration}.

One of the most common ways to evaluate calibration is through reliability diagrams, which compare predicted confidence levels with observed accuracy (see~\figref{fig:example_diagram}(top)). Interestingly, in communication systems, the failure probabilities that users care about can span several orders of magnitude, from rare events to more frequent occurrences, making standard calibration visualizations on linear scales inadequate. To address this, logarithmic binning~\cite{10757632} provides a more effective way to assess calibration across broad probability ranges, offering a clearer picture of how well NNs perform at low probability events.

Improving calibration is particularly critical in ML-driven wireless networks, where miscalibrated models can result in inefficient resource allocation, increased failure rates, or unnecessary precautionary measures.
This issue becomes even more significant in ultra-reliable low-latency communications (URLLC) applications, such as industry automation, where accurately predicting rare outage events is essential for maintaining safety and system efficiency. Motivated by these concerns, we propose applying post-processing calibration methods to improve the reliability of an ML-based outage predictor. We examine how its calibration properties influence resource allocation errors in a single-user, multi-resource allocation framework. In particular, we focus on the classification threshold of this predictor, which determines how calibrated confidence scores are used to predict outages and helps achieve a target system outage probability (OP), which serves as our metric for resource allocation error. 

\subsection{Related Work}
\subsubsection{ML for Wireless Systems}

ML has become an essential tool for ensuring reliable connectivity and efficient resource allocation in dynamic wireless networks. In particular, deep learning (DL) has been widely applied to tasks such as blockage prediction and link quality assessment~\cite{moon2022online,9562750,wu2022deep,charan2021vision,alrabeiah2020deep}. For example, in~\cite{moon2022online}, a deep NN was trained to map user positions and traffic demands to blockage status and optimal beam index, achieving a blockage prediction accuracy of 90\% in millimeter wave (mmWave) systems. The authors of~\cite{9562750} developed a meta-learning based framework for mmWave and terahertz systems, where a recurrent NN was trained using a few data samples to predict potential blockages. In~\cite{wu2022deep}, models were trained on mmWave received power sequences to identify pre-blockage signatures. In~\cite{charan2021vision}, beam sequence data was processed using a gated recurrent unit network to forecast line-of-sight blockages. Additionally,~\cite{alrabeiah2020deep} proposed a NN that learned to predict both the optimal mmWave beam and the blockage status using sub-6 GHz channel information.

Beyond blockage prediction, DL has also been employed for power allocation in massive multiple-input multiple-output (MIMO) systems~\cite{sanguinetti2018deep} and channel estimation~\cite{8682819,8761432,8761934,9795301}. For instance,~\cite{sanguinetti2018deep} demonstrated a significant reduction in the complexity and processing time of the optimization process for massive MIMO systems;~\cite{8682819} used a convolutional NN for wideband channel estimation in mmWave massive MIMO systems;~\cite{8761432} and~\cite{8761934} used long short-term memory (LSTM) networks to make channel predictions in body area networks and future signal strength, respectively; and~\cite{9795301} explored DL techniques to predict future received signal strength variations in device-to-device (D2D) communications channels. DL has also found extensive use in traffic prediction. For example,~\cite{8057090} proposed a hybrid DL model, while~\cite{8264694} applied a recurrent NN for spatiotemporal traffic prediction in cellular networks;~\cite{9014115} compared the performance of statistic learning and DL in user traffic prediction; and~\cite{8667446} developed a DL architecture and applied transfer learning among different scenarios for traffic prediction.

Further expanding its utility, numerous studies have used ML techniques, such as deep reinforcement learning (DRL), for a wide range of wireless communication tasks, including mobile networking~\cite{pmlr-v97-jay19a,8845211,8422824}, spectrum sharing in cognitive radio networks~\cite{li2018intelligent}, scheduler design~\cite{8425580,meng2019delay,8100645}, resource allocation~\cite{10705105,8468000,ye2019deep,8698845,hua2019gan,7997286,8377343,8540003,8736846}, and energy efficiency optimization in D2D enabled heterogeneous networks~\cite{zhang2020energy}. For example,~\cite{pmlr-v97-jay19a} addressed internet congestion by dynamically adjusting sender data rates, while~\cite{8425580} applied DRL to select scheduling policies that minimize packet delays and drop rates. In~\cite{10705105}, the authors proposed an optimal classifier for an ML-assisted resource allocation system,~\cite{8468000} minimized long-term system power consumption in Fog radio access network, and~\cite{ye2019deep} proposed a decentralized DRL-based framework for resource allocation in vehicle-to-vehicle communications, effectively supporting both unicast and broadcast modes. Similarly,~\cite{8698845} minimized average latency in mobile edge computing (MEC) Internet-of-Things networks. In a related context, approaches such as Quantum DRL have also been used to support joint resource allocation and task offloading in MEC~\cite{10752359} and to enhance computational learning speed~\cite{10318071}.
Additionally, several studies have proposed ML-driven strategies for resource allocation in URLLC systems\cite{8491089, 9013851}, as well as for managing the coexistence of URLLC and enhanced mobile broadband services in 5G New Radio networks~\cite{alsenwi2021intelligent, hsu2022embb, huang2020deep}. 

\subsubsection{Custom Loss Functions for ML-Driven Wireless Systems}
Most ML-based communication models highlighted previously rely on traditional loss functions, such as binary cross-entropy (BCE) and mean squared error (MSE), to optimize model performance. 
While these loss functions are effective for general classification and regression tasks, they do not always align with the specific requirements of wireless networks~\cite{10624794}. In particular, conventional training objectives often fail to capture communication-specific constraints, such as OPs, resource efficiency, and latency guarantees. This misalignment can lead to suboptimal performance, prompting research into more specialized optimization approaches.

To address these limitations, custom loss functions have been designed to incorporate domain knowledge and better reflect communication system objectives~\cite{9369424}. Such loss functions have been successfully applied in hybrid beamforming for MIMO systems~\cite{9130130}, optimizing reflection coefficients in reconfigurable intelligent surfaces~\cite{10166830}, and resource allocation in cell-free massive MIMO networks~\cite{10094043} and URLLC~\cite{9013851}. In addition,~\cite{10443669} introduced the outage loss function (OLF) to specifically minimize OPs during model training. Other applications include wireless channel estimation, where they preserve low-rank properties of the estimated channel matrix~\cite{9252921}, cellular traffic prediction~\cite{8932440}, received signal strength indicator (RSSI)-based wireless indoor localization~\cite{10844973}, and statistical federated learning for beyond 5G network slicing~\cite{9534716}. 

\subsubsection{ML Calibration}
Regardless of the loss function used, calibration remains a key challenge in ML-driven communication systems. 
To address this issue, Bayesian learning~\cite{MAL-052,Simeone_2022} models epistemic uncertainty through parameter distributions to improve probabilistic predictions. It has been applied to a range of communication tasks, including network resource allocation~\cite{9430561,9540910,8251223}, massive MIMO detection~\cite{9712258,9953978,9536454}, and channel estimation~\cite{9448139}. While exact Bayesian learning offers formal calibration guarantees, these rely on strong assumptions. Specifically, they assume that the model is well-specified, meaning the NN has enough capacity to represent the ground-truth data generation process, and that its predictive distribution for continuous outputs is unimodal. These assumptions may not be satisfied in cases where sufficient data is not available. In~\cite{10081084}, the authors explored the application of robust Bayesian learning to wireless communication systems, focusing on applications such as automated modulation classification, RSSI-based localization, as well as channel modeling and simulation. However, this approach does not offer formal calibration guarantees. Conformal prediction~\cite{JMLR:v9:shafer08a} is a general framework for constructing set predictors that offer formal, distribution-free calibration guarantees and has been applied to problems such as symbol demodulation and modulation classification~\cite{10262367}. More recently, it has been explored as a tool to support uncertainty quantification through pre-deployment calibration, online monitoring, and post-deployment diagnostic analysis~\cite{simeone2025conformalcalibrationensuringreliability}.
Another widely used approach is post-processing calibration, which adjusts confidence scores using a validation set. Common techniques include Platt scaling, isotonic regression, beta calibration for binary classification, and temperature or Dirichlet scaling for multiclass settings~\cite{6274982,zadrozny2002transforming,kull2017beta,kull2019beyond,9022283,gupta2021distributionfreecalibrationguaranteeshistogram,9710154}.
\subsection{Contributions}
Our contribution focuses on studying the calibration of the outage predictor proposed in~\cite{10443669} for an ML-assisted resource allocation system, providing a more precise understanding of its role in resource allocation. Specifically, we summarize our key contributions as follows: 
\begin{enumerate}  
\item We show that for perfectly calibrated models, the system OP—in the infinite-resource limit—equals the model’s expected confidence, conditioned on it being less than its classification threshold, and equals the resource’s underlying OP when only a single resource is available.
\item We conduct a rigorous simulation-based analysis in a Rayleigh fading environment using post-processing calibration techniques, demonstrating our theoretical finding that well-calibrated models enable straightforward classification threshold selection. For instance, for a well-calibrated model with a sufficiently large number of resources, system designers need only choose a classification threshold equal to their desired OP, resulting in a system OP that is lower than the selected threshold.
\item We show that well-calibrated models form a special case of a more general class of models that necessarily improves OP. Specifically, we establish a monotonicity condition that accuracy-confidence functions must satisfy in order to improve system OP.
\item We show that post-processing calibration cannot improve the system’s minimum achievable OP. This follows from the~\textit{data processing inequality}, which ensures that such calibration methods do not introduce additional information about future channel states.
\item Finally, we present a preliminary proof of concept demonstrating the application of the resource allocation technique and the OLF used in this work, and in~\cite{github}. This demonstration is intended for a broader audience.
\end{enumerate}
This work builds on~\cite{11140554} and presents a significantly more comprehensive study of the outage predictor calibration within this system. A small part of the simulation work
established here, appears in~\cite{raina2025ml}. In this paper, we analyze how calibration affects the classification
threshold, enabling more effective resource selection, while also incorporating the additional contributions outlined
above. For reproducibility purposes, we have
made our code publicly available\footnote{{https://github.com/ML4Comms/greedy-resource-allocation-outage-classification}~\cite{github}.}.

The remainder of this paper is organized as follows: Section~\ref{sec:system_model} revisits the resource allocation strategy incorporating an outage predictor, as proposed in~\cite{10443669}. It then introduces the OLF used for its training, before providing an overview of calibration concepts. Section~\ref{sec: on_the_impact_of_calibration} discusses the key theoretical results of this work. Section~\ref{sec:methodology} begins with a description of the data generation process, outlines the essential metrics for evaluating calibration performance, and provides an overview of the post-processing calibration techniques implemented in this paper. Section~\ref{sec:results} presents the numerical results that demonstrate the findings established in Section~\ref{sec: on_the_impact_of_calibration}, and discusses the proof of concept for the resource allocation technique introduced in Section~\ref{sec:system_model}. Finally, Section~\ref{sec:conclusion} concludes the paper with some finishing remarks.
\section{System Model}
\label{sec:system_model}
In this section, we revisit the single-user, multi-resource system model from~\cite{10443669} and restate the essential formulations to provide the necessary context for this study. We focus on the ML-based outage predictor utilized by the user and examine how its calibration properties influence the resource allocation framework.
\subsection{Channel Model for a Resource}
Each resource $j \in \mathcal{R}$ has a fluctuating channel state $h_j(t)\in\mathbb{C}$, with channel states assumed to be independent and identically distributed (i.i.d.) across distinct resources. Furthermore, the correlation between $h_j(t)$ and $h_j(t + l)$ decreases as $l \to \infty$. The user is equipped with an ML-based outage predictor that assists in selecting the appropriate resource by learning these temporal correlation patterns. For $k, l \in \mathbb{N}$, the vectors of past and future channel states at time $t$ are given by $\accentset{\leftarrow}{H}_{j}^k{(t)} \triangleq \left[h_{j}{(t-k+1)},\cdots, h_{j}{(t)} \right]^{T}$, and $\accentset{\rightarrow}{H}_{j}^l(t) \triangleq \left[h_j(t+1), \cdots, h_j(t+l)\right]^T$, respectively.


The capacity $C\left({\accentset{\rightarrow}{H}_{j}^l}{(t)}\right) \in \mathbb{R}^+$ measures resource $j$'s ability to support communication from time $t+\!1$ to $t+\!l$. For a quasi-static Gaussian channel, it is given by~\cite[eq. (5.80)]{10.5555/1111206}:

\begin{equation}
\!C\!\left(\!{\accentset{\rightarrow}{H}_{j}^l}{(t)}\!\right)\! =  \!\sum_{i=1}^{l}\log_2\left( 1 + \mathtt{SNR}\left| h_j\left(t + i\right)\right|^2\right)~~~~\rm{\!bits\!/s/\!Hz},\label{eq:Gaussian_channel_capacity}
\end{equation}
where $\mathtt{SNR}$ is the average signal-to-noise ratio per sample. 
If the user’s requirements, defined by a key performance indicator (KPI) such as transmission rate, exceed this capacity, the resource is considered to be in outage; otherwise, it is deemed satisfactory. 
The required KPI level is specified by a rate threshold, $\gamma_{th}$. 

The OP for a single resource $j$ can be expressed as
\begin{equation}
P_j (\gamma_{th}) = \mathbb{P}  \left[C\left({\accentset{\rightarrow}{H}_{j}^l}{(t)}\right)< \gamma_{th}\right],
\label{eq:outage_probability_for_singe_resource}
\end{equation} where $\mathbb{P}[\cdot]$ represents the ground-truth distribution of the observation state sequences $\accentset{\rightarrow}{H}_{j}^l{(t)}$.

\subsection{ML-based Resource Allocation}
This work focuses on a calibrated ML-based predictor, unlike~\cite{10443669}, which considers the general case where the model may or may not be well-calibrated.
The predictor's output $Q\left(\accentset{\leftarrow}{H}_{j}^k{(t)};\Theta\right)\in [0,1]$, determined by the parameters \( \Theta \), provides an estimate of the conditional probability:
\[
\mathbb{P} \left[ C\left(\accentset{\rightarrow}{H}_{j}^l{(t)}\right) < \gamma_{th} \mid \accentset{\leftarrow}{H}_{j}^k{(t)}\right],
\]
where the conditioning is over the information $\accentset{\leftarrow}{H}_{j}^k{(t)}$ available at the time of prediction. An outage is predicted if $Q\!\left(\!\accentset{\leftarrow}{H}_{j}^k{(t)};\Theta\right)\! > \!q_{th}$,
where $q_{th}$ is the predictor's classification threshold.

The user relies on an ML-based outage predictor for resource allocation, where resources are indexed from $j=1$ to $|\mathcal{R}|$. 
Each time the predictor predicts an outage event, the index is incremented by 1. The predictor uses the input vector $\accentset{\leftarrow}{H}_{j}^k{(t)}$ for each resource \( j \), producing an output in \([0,1]\) that classifies whether the upcoming \( l \) channel samples $\accentset{\rightarrow}{H}_{j}^l{(t)}$ will support outage-free communication. If no resource meets this condition, the model selects the final resource.

In this study, calibration is defined as the ML predictor’s confidence matching the true outage frequency, with perfect calibration achieved when predictions with confidence $q$ correspond to an actual outage frequency of $q$.
In order to quantify the calibration quality of the predictor, we adopt the standard characterization in terms of \textit{reliability diagrams}. A reliability diagram is a visual representation of model calibration, plotting the observed accuracy as a function of confidence~\cite{guo2017calibration}. In this context, the accuracy-confidence function
$A_j(q)$ describes the probability of an outage
event given that the predictor provides an outage confidence of $q$. For our outage predictor $Q\left(\accentset{\leftarrow}{H}_{j}^k{(t)}\right)$, the accuracy-confidence function is formally defined as
\begin{equation}
A_j(q) \triangleq \mathbb{P} \left[ C\left(\accentset{\rightarrow}{H}_{j}^l{(t)}\right) < \gamma_{th} \,\middle|\, Q\left(\accentset{\leftarrow}{H}_{j}^k{(t)};\Theta\right) = q \right].
\label{eq:accuracy_function}
\end{equation}

\begin{figure}[t]   \centering    \begin{minipage}{0.46\textwidth}        \centering        \includegraphics[width=\textwidth]{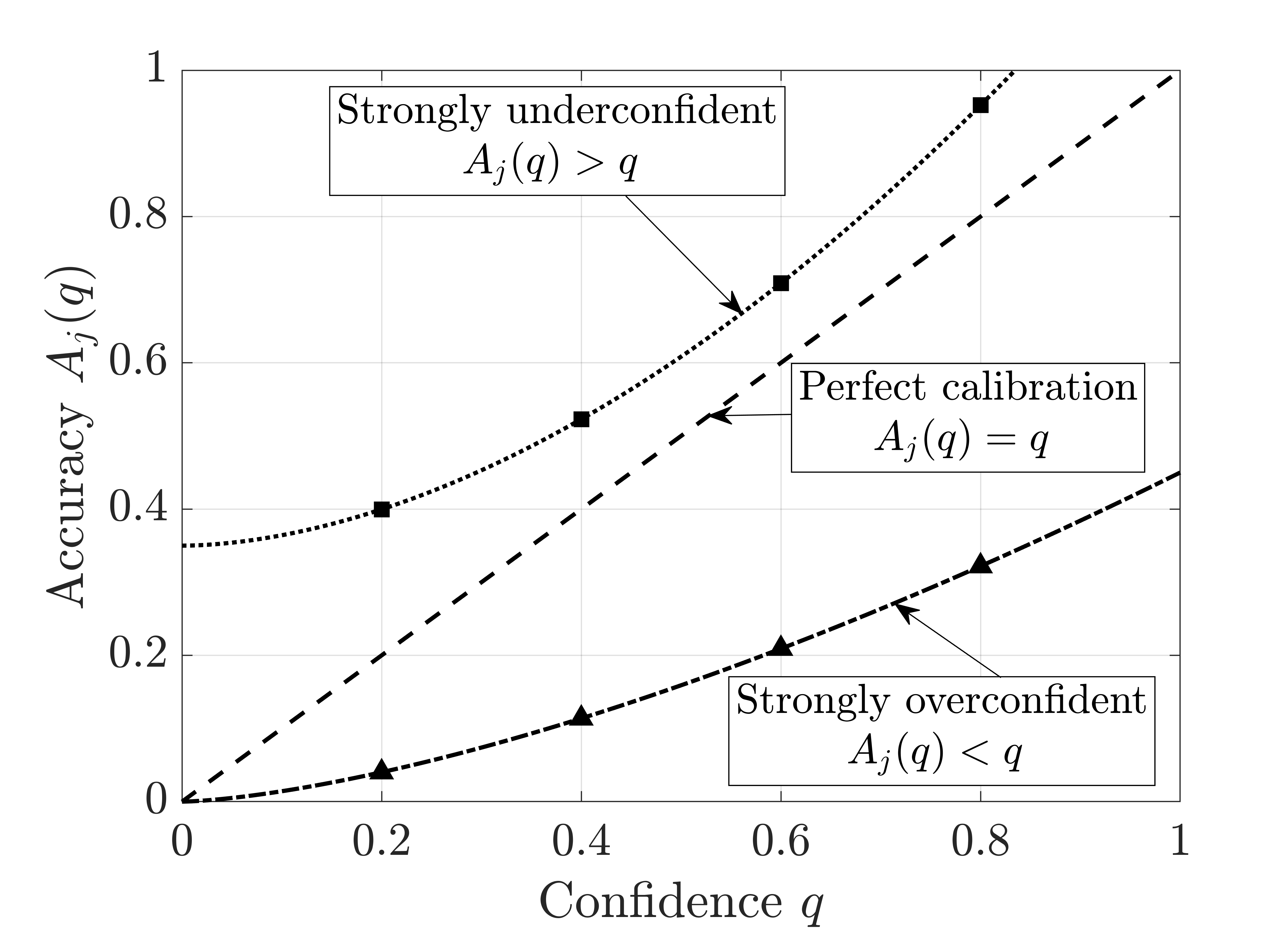}  \end{minipage}  
 \begin{minipage}{0.46\textwidth}       \centering       \includegraphics[width=\textwidth]{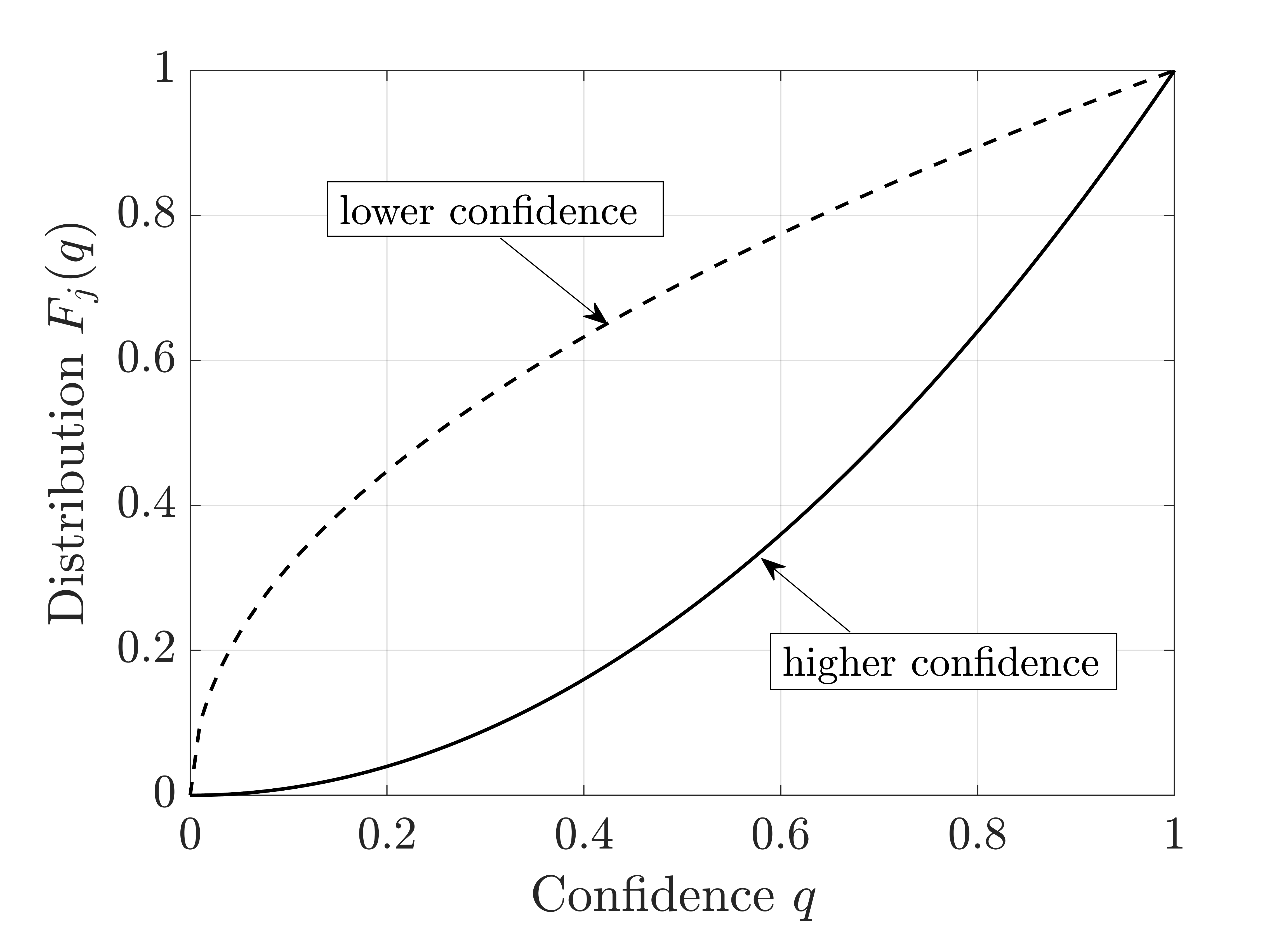}  \end{minipage}  \caption{Reliability diagram showing the accuracy-confidence function $A_j(q)$ (top), and the confidence distribution $F_j(q)$ (bottom).}   \label{fig:example_diagram}\end{figure}

The accuracy-confidence function gauges the degree to which a predictor’s confidence level matches the true predictive accuracy for any fixed confidence level $q$. This does not provide information on how frequently the predictor returns a confidence level $q$. That information is given by the confidence distribution $F_j(q)$, defined as the cumulative distribution function
\begin{equation}
F_j(q) = \mathbb{P}\left[Q\left(\accentset{\leftarrow}{H}_{j}^k{(t)}\right) \leq q\right],
\label{eq:conf_distribution}
\end{equation}
which gives the fraction of past-sample histories 
$\accentset{\leftarrow}{H}_{j}^k{(t)}$
for which the predictor returns a confidence level less than or equal to $q$.


Illustrative plots of the functions $A_{j}(q)$ and $F_{j}(q)$ are shown in~\figref{fig:example_diagram}.
The top figure shows three accuracy-confidence curves corresponding to the extreme cases: perfect calibration $\left(A_j(q)=q\right)$, strong under-confidence $\left(A_j(q) > q\right)$, and strong over-confidence $\left(A_j(q) < q\right)$, for all $q\in [0,1]$. The bottom figure shows two examples of confidence distributions, where a lower $F_{j}(q)$ indicates a model that tends to produce a larger fraction of confident decisions.

Since the predictor is generally incorrectly calibrated, we have the non-equality
\begin{equation}
    A_j(q) \neq q
\end{equation}

The goal of this paper is to study the effect of applying calibration strategies to improve the calibration properties of predictive models, so that
\begin{equation}
    A_j(q) \approx q.
\end{equation}

\noindent Let $Q^\star \in \{ Q, Q^c \}$, where $Q^c$ denotes the calibrated predictor. Similar to~\cite{guo2017calibration}, we assume that the calibrated predictor be a post-processing of the original scores:
\begin{equation}
    Q^c \;=\; \mathcal C\bigl(Q;\mathcal D\bigr)
\;=\;
r\circ Q,
\label{eq:Qc = rQ}
\end{equation}
where
the recalibration map
\begin{equation}
    r : [0,1] \;\longrightarrow\; [0,1]
\end{equation}
is learned solely from a held-out calibration dataset \(\mathcal D\). Here, $\mathcal C\bigl(Q;\mathcal D\bigr)$ represents the calibration procedure, and $r\circ Q$ represents function composition, i.e., the calibrated predictor is obtained by applying 
$r$ to the output of $Q$.
Since $Q^c\!\left(\accentset{\leftarrow}{H}_{j}^k{(t)}\right)$ is obtained by (possibly stochastic) processing of $Q\!\left(\accentset{\leftarrow}{H}_{j}^k{(t)}\right)$, the data processing inequality implies
\noindent
\begin{equation}
\!I\!\left(\!C\!\left(\!\accentset{\rightarrow}{H}_{j}^l{(t)}\!\right);\!Q^c\!\left(\accentset{\leftarrow}{H}_{j}^k{(t)}\!\right)\!\right)\;\!\!\le\;I\!\left(C\left(\accentset{\rightarrow}{H}_{j}^l{(t)}\!\right);\!Q\!\left(\accentset{\leftarrow}{H}_{j}^k{(t)}\!\right)\!\right)\!,
\label{eq:DPI}
\end{equation}
where $I(\cdot,\cdot)$ is the mutual information~\cite{cover1999elements}. This ensures that no additional information about \(C\left(\accentset{\rightarrow}{H}_{j}^l{(t)}\right)\) is injected during calibration. Next, we describe the resource allocation strategy used in this study.

\subsection{Greedy Resource Allocation}
We adopt the resource allocation strategy from~\cite{10443669}, in which the user scans the available resources $\mathcal{R}$ and selects a resource $j'$ based on whether the outage predictor $Q^\star\left(\accentset{\leftarrow}{H}_{j}^k{(t)}\right)$ indicates successful communication. The predictor sequentially scans resources $\mathcal{R} = \{1, 2, \dots, |\mathcal{R}|\}$, stopping once it predicts successful communication for a given resource. If no resource meets this condition, the final resource is selected.


Formally, let
\begin{equation}
   \tilde{\mathcal{R}}({q_{th}}) = \left\{ j \in \mathcal{R}~\text{s.t.}~
Q^\star\left(\accentset{\leftarrow}{H}_{j}^k{(t)}; \Theta\right) \leq q_{th} \right\}
,
\end{equation}
represent the subset of resources where the outage predictor predicts no outages. Then the greedy allocation scheme selects
\begin{equation}
    j'(q_{th}) = 
    \begin{cases} 
        \min_{j \in \tilde{\mathcal{R}}(q_{th})}~j~~& \text{if } \tilde{\mathcal{R}}({q_{th}}) \neq \emptyset, \\
        |\mathcal{R}| & \text{otherwise}.
    \end{cases}
   \label{eq:rgreedy1}
\end{equation}

Let
$P^\star_{|\mathcal{R}|}(\gamma_{th}, q_{th}) \in \{ P_{|\mathcal{R}|}(\gamma_{th}, q_{th}),\; P^c_{|\mathcal{R}|}(\gamma_{th}, q_{th}) \}$ denote the OP of a system with $|\mathcal{R}|$ resources, when using either the uncalibrated predictor $Q$  or the calibrated predictor $Q^c$, respectively. These are formally defined as:
\begin{equation}
P^\star_{\vert\mathcal{R}\vert} (\gamma_{th}, q_{th}) \triangleq \mathbb{P} \left[C\left(\accentset{\rightarrow}{H}_{j'(q_{th})}^{l}(t)\right) < \gamma_{th} \right],    \label{eq:OP}
\end{equation}
where $j'(q_{th})$ is the selected resource index according to~\eqref{eq:rgreedy1}.
From~\cite[eq. (13)]{10443669}, the OP in~\eqref{eq:OP} can be expressed as:
\noindent\begin{align}
P^\star_{|\mathcal{R}|}(\gamma_{th}, q_{th})
&= {P_j(\gamma_{th}) \left(1 - F_{Q^\star}(q_{th})\right)^{|\mathcal{R}|-1}} \nonumber\\
&\quad + \!{P^\star_\infty(\gamma_{th}, q_{th})
       \!\left(1 -\! \left(1 - \!F_{Q^\star}(q_{th})\right)^{|\mathcal{R}|-1}\!\right)},
\label{eq:main_outage_expression}
\end{align}
where $F_{Q^\star}(q_{th})$ is the cumulative distribution function of the predictor’s output $Q^\star$, and $P^\star_\infty(\gamma_{th}, q_{th})$ is the OP of a system with an infinite number of resources, given by:
\noindent \begin{equation}
\!\!\!P^\star_{\infty}(\gamma_{th}, q_{th}) \!= \mathbb{P} \!\left[ C\left(\accentset{\rightarrow}{H}_{j}^{l}(t)\!\right)\! < \gamma_{th} \middle| Q^\star\!\left(\accentset{\leftarrow}{H}_{j}^{k}(t)\!\right) \le q_{th} \right]\!,
\label{eq:OP_infinite_general}
\end{equation} 
with $Q^\star \in \{ Q, Q^c \}$ as before.

Using~\eqref{eq:main_outage_expression}, and the law of total probability, the OP can also be expressed in terms of the predictor's confidence distribution function and the accuracy-confidence function. From~\eqref{eq:main_outage_expression}, the term $(1 - F_{Q^\star}(q_{th}))^{|\mathcal{R}| - 1}$ represents the probability that none of the resources meets the condition $Q^\star\left(\!\accentset{\leftarrow}{H}_{j}^k{(t)}\right) \!\leq q_{th}$, in which case the final resource is selected, and the OP is given by $P_j(\gamma_{th})$. 
Conversely, the term $1 - (1 - F_{Q^\star}(q_{th}))^{|\mathcal{R}| - 1}$ represents the probability that at least one of the resources satisfies the condition $Q^\star\left(\!\accentset{\leftarrow}{H}_{j}^k{(t)}\right) \!\leq q_{th}$. In this case, the OP is given by the expected accuracy $\mathbb{E}\left[A_j\left(Q^\star\left(\accentset{\leftarrow}{H}_{j}^k(t)\right)\right) \,\middle|\, Q^\star\left(\accentset{\leftarrow}{H}_{j}^k(t)\right) \leq q_{th} \right]$. Note that, due to the i.i.d. assumption on channel states across resources, the confidence distribution $F_j(q_{th})$ is identical for all $j \in \mathcal{R}$. Therefore, $F_{Q^\star}(q_{th}) \equiv F_j(q_{th})$ for any resource $j$, and we use the notation $F_j(q_{th})$ in what follows. Thus, the OP is given by:
\begin{align}
 P^\star_{|\mathcal{R}|}(\gamma_{th}, q_{th}) &= P_j(\gamma_{th})\left(1 - F_j(q_{th})\right)^{|\mathcal{R}|-1} \nonumber\\&+ \!\mathbb{E}\!\left[A_j\left(Q^\star\!\left(\accentset{\leftarrow}{H}_{j}^k{(t)}\right)\right) \middle| Q^\star\left(\!\accentset{\leftarrow}{H}_{j}^k{(t)}\right) \!\leq q_{th}\right]
 \nonumber\\ &\times \left(1 - (1 - F_j(q_{th}))^{|\mathcal{R}|-1}\right).
 \label{eq:OP_case1_greedy_scheme} 
\end{align}
Equation~\eqref{eq:OP_case1_greedy_scheme} captures the relationship between the system's OP, the accuracy-confidence function, and the confidence distribution function. This equation is further used in Theorem~3 (see Appendix~\ref{app:thm2_proof}), which established a monotonicity condition that the accuracy-confidence function must satisfy in order to improve system OP.

Furthermore, we say that a predictor is useful at a threshold level $q_{th}$ if the inequality
\begin{equation}
    P^\star_{\vert\mathcal{R}\vert} (\gamma_{th}, q_{th}) < \underset{j\in\mathcal{{R}}}{\text{min}}~P_j
    \label{eq:OP_max}
\end{equation}
holds. This indicates that the predictor allows the selection of a resource~\eqref{eq:rgreedy1} that supports a probability of outage lower than the minimum OP $\min_{j \in \mathcal{R}}~P_j$ achievable through random selection of a resource.

The OLF, customized for this resource allocation system, is defined in~\cite[eq. (29)]{10443669}. This loss function is utilized to train the outage predictor, enabling the generation of the results presented in Section~\ref{sec:results}. For brevity, we drop the dependence of the sequences $\accentset{\leftarrow}{H}_{j}^k{(t)}$ and $\accentset{\rightarrow}{H}_{j}^l{(t)}$ on $t$, as well as the dependence on $j$ for $P \triangleq P_{j}$, accuracy-confidence function $A(q) \triangleq A_j(q)$, and confidence distribution $F(q) \triangleq F_j(q)$. Additionally, where appropriate, we drop the dependence of the predictor's confidence $Q^\star(\cdot)$ and of $C(\cdot)$ on $\accentset{\leftarrow}{H}_{j}^k{(t)}$ and $\accentset{\rightarrow}{H}_{j}^l{(t)}$, respectively. We denote by $Q^\star$ a random variable with the same distribution as any confidence level $Q^\star$ assigned by the predictor to a resource $j$. 

Before moving on to the next section, we reiterate our problem statement here. This paper builds on~\cite{11140554} and substantially expands on it by providing a comprehensive analysis of the calibration behavior of the outage predictor in an ML-assisted resource allocation system. In particular,~\cite{11140554} analyzed the general case where the outage predictor may or may not be well calibrated. In such cases, the predictor’s confidence scores may not accurately reflect the true probability of outages. Central to our study is the characterization of the system's OP under perfect calibration, which helps guide the choice of the classification threshold required to achieve a target OP when sufficient resources are available. We then apply post-processing calibration techniques to validate our theoretical findings.
\section{On the Impact of Calibration}
\label{sec: on_the_impact_of_calibration}
In this section, we present the main results of this work, detailing how the output of the calibrated predictor relates to the system's outage performance metrics.

\subsection{Main Results}
Our analysis reveals that perfectly calibrated predictors satisfy key properties related to the system's OP. Specifically, for $|\mathcal{R}|\to \infty$, the OP equals the expected confidence conditioned on $Q^c \leq q_{th}$; in the single resource case, it equals the overall expected confidence. These results help identify an appropriate $q_{th}$ to meet target performance results, providing guidance for designing wireless systems with specific reliability levels. This is captured in the following theorem.
\begin{thm}
\label{thm thm1}
Consider the resource allocation system described in Section~\ref{sec:system_model}. Then, for a perfectly calibrated predictor, we have:
\begin{align}
    P &= \mathbb{E}[Q^c], \label{eq:P_equals_EQ} \\
    \text{and} \qquad P^c_{\infty}(\gamma_{th}, q_{th}) &= \mathbb{E}[Q^c \mid Q^c \leq q_{th}], \label{eq:Pinf_equal_Eqth_thm1}
\end{align}
\noindent
so that we can express:
\begin{align}
    \noindent \!\!P^c_{|\mathcal{R}|}(\gamma_{th}, q_{th}) \!&= \mathbb{E}[Q^c](1 - \!F(q_{th}))^{|\mathcal{R}|-1} \notag \\
    &\quad + \mathbb{E}[Q^c \!\mid Q^c\!\leq q_{th}]\!\left(\!1 - \!(1 - \!F(q_{th}))^{|\mathcal{R}|-1}\!\right). \label{eq:PR_expression}
\end{align}
\begin{IEEEproof} See Appendix~\ref{app:thm1_proof}. 
\end{IEEEproof}
\end{thm}
Theorem~\ref{thm thm1} reveals that for a perfectly calibrated predictor, the OP is determined by the predictor’s expected confidence, either conditioned on being below $q_{th}$ (in the infinite-resource case) or across all predictions (in the single-resource case). Moreover,~\eqref{eq:Pinf_equal_Eqth_thm1} tells us that when the number of resources is sufficiently large, one should select a value of $q_{th}$ such that the conditional expected OP of the model is equal to their desired OP.
Using Theorem~\ref{thm thm1}, we can further establish simpler results on the relationship between the system's OP and the classification threshold $q_{th}$ in the well-calibrated scenario.
\begin{corollary}
\label{cor:pinf_bound}
For a perfectly calibrated predictor, the OP satisfies:
\begin{equation}
   P^c_{\infty}(\gamma_{th}, q_{th}) \leq q_{th},
   \label{eq:Pinf_leq_qth}
\end{equation}
\begin{align}    P^c_{\vert\mathcal{R}\vert}(\gamma_{th}, q_{th}) & \leq \mathbb{E}[Q^c] (1 - F(q_{th}))^{|\mathcal{R}| - 1}\nonumber\\
    & + q_{th}\left(1 - (1 - F(q_{th}))^{|\mathcal{R}|-1}\right).
     \label{eq:Theorem1_eqn2}
\end{align}
\end{corollary}
This corollary further simplifies the process of selecting an appropriate $q_{th}$. Equation~\eqref{eq:Pinf_leq_qth} tells us that if the model is well-calibrated, all one needs to do is select a $q_{th}$ equal to their desired OP, and provided there are sufficient resources, they should achieve a system OP that is better than their desired OP. For example, if a system requires an OP of 0.01, setting $q_{th}= 0.01$ ensures that the system OP will be better than the desired OP.


Note that although calibration improves the alignment between predicted confidence and actual accuracy, it does not increase the informativeness of the predictor. Post-processing calibration methods adjust the output scores using a held-out validation set, but they do not alter the underlying model or incorporate any new information. As a result, such methods cannot reduce the system’s minimum achievable OP, in accordance with the data processing inequality, as formalized in the following theorem.
\begin{thm}
\label{thm thm3}
For any post-processing calibration procedure satisfying~\eqref{eq:Qc = rQ}, applying this calibration will not improve the system's OP performance. i.e.,
\begin{equation}
    \min_{q_{th}} P^c_{\vert\mathcal{R}\vert}(\gamma_{th}, q_{th}) \geq \min_{q_{th}}P_{\vert\mathcal{R}\vert}(\gamma_{th}, q_{th}).
\end{equation}
    \begin{IEEEproof}        
    See Appendix~\ref{app:thm3_proof}.
\end{IEEEproof}\end{thm}

\begin{figure}[!t]    \centering\includegraphics[trim= 1.0cm 15.3cm 1.9cm 0.20cm, clip, width=\textwidth]{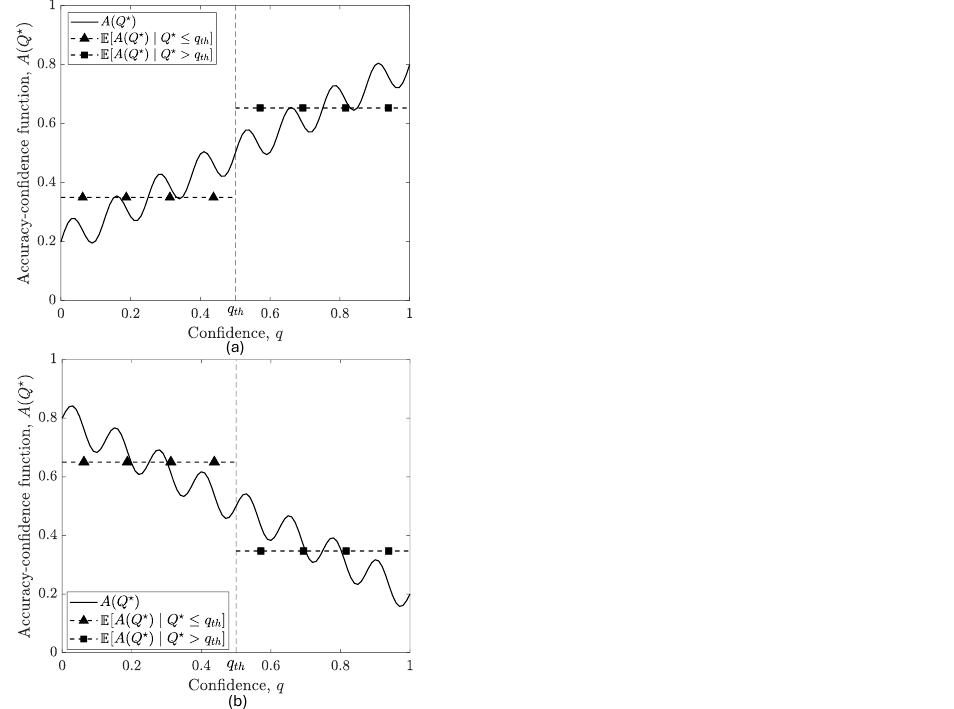}  \caption{Illustration of Theorem~\ref{thm thm2}, where (a) a predictor is useful at $q_{th} = 0.5$, satisfying~\eqref{eq:useful_predictor}, and (b) a predictor is not useful at the same $q_{th}$.}
\label{fig:thm2_fig}
\end{figure}
\vspace{-1pt}
We now come to the final theoretical result of our work, which introduces the concept of monotonicity that the accuracy-confidence function must satisfy for the predictor to be useful (i.e., satisfying~\eqref{eq:OP_max}).
\begin{thm}
\label{thm thm2}
Given a threshold level $q_{th}$, a predictor is useful (i.e., it satisfies~\eqref{eq:OP_max}) if the inequality $\mathbb{E}\left[A(Q^\star)| Q^\star \leq q_{th}\right]$~$<$~$P$ is satisfied. Furthermore, $\mathbb{E}\left[A(Q^\star)| Q^\star \leq q_{th}\right] < P$ is satisfied if and only if
    \begin{equation}\mathbb{E}\left[A(Q^\star)| Q^\star \leq q_{th}\right] <  \mathbb{E}\left[A(Q^\star)| Q^\star >q_{th}\right].        \label{eq:useful_predictor}    \end{equation}   
Moreover, at the critical point defined by: \begin{equation}     \mathbb{E}[A(Q^\star) \mid Q^\star \leq q_{th}] = \mathbb{E}[A(Q^\star) \mid Q^\star > q_{th}]\end{equation} there is no net change in the system's outage performance.
    \begin{IEEEproof}        
    See Appendix~\ref{app:thm2_proof}.
    \end{IEEEproof}\end{thm}
This theorem shows us that the conditional expectation of the predictor's accuracy-confidence function should be smaller when conditioned on values less than or equal to $q_{th}$ than when conditioned on values greater than $q_{th}$. 
A trivial class of accuracy-confidence functions that achieve this are those that are monotonically increasing. As an example, if $A(q) = q$ for all values of $q$, as is the case for perfectly calibrated predictors, then $A$ is a monotonically increasing function and so the corresponding predictor necessarily improves the OP. There may be examples of strongly over-confident and strongly under-confident predictors that also satisfy this monotonicity property  (see Fig.~\ref{fig:example_diagram}~(top)). Importantly, $A(q)$ being monotonically increasing is a sufficient but not necessary condition required for a predictor to be useful. For example, it is possible for $A(q)$ to be non-monotonic in the regions $[0, q_{th})$ and $[q_{th}, 1]$ whilst also satisfying \( A(q) > A(q') \) for \( q' \in [0, q_{th}) \) and \( q \in [q_{th}, 1] \). Such a function would also satisfy~\eqref{eq:useful_predictor} (see Fig.~\ref{fig:thm2_fig}~(a)). Conversely, we can also see that if a predictor’s accuracy-confidence function is monotonically decreasing, then it will necessarily hinder the system’s OP (see Fig.~\ref{fig:thm2_fig}~(b)). Again, using a similar example as above, this is a sufficient but not necessary property of a predictor that hinders performance.

Furthermore, this theorem tells us that at a critical point, defined by the condition that the average accuracy for instances where $Q^\star \leq q_{th}$ equals that for $Q^\star > q_{th}$, the system’s outage performance exhibits no overall change.

\section{Methodology}
\label{sec:methodology}
This section describes the data-generation process and outlines the key metrics used to evaluate the calibration performance of the outage predictor. It also provides an overview of the post-processing calibration techniques implemented in this study, along with relevant implementation details.
\begin{table}[b]
\small
\centering
\renewcommand{\arraystretch}{1.1}
\caption{Model Architectures and Hyperparameters}
\label{table1}
\begin{tabular}{|p{3.36cm}|p{4.7cm}|}
\hline
\textbf{Component} & \textbf{Specification} \\
\hline
\multicolumn{2}{|c|}{\textbf{Architecture (common to both models)}} \\
\hline
Layer 1: LSTM layer & 1 layer with 32 hidden units \\
Layer 2: Dense layer & 10 units with PReLU activation \\
Layer 3: Dense layer & 1 unit with sigmoid activation \\
\hline
\multicolumn{2}{|c|}{\textbf{Training Hyperparameters}} \\
\hline
Learning rate & 0.001 \\
Number of epochs & 30 \\
Epoch size & 150 \\
Batch size & Number of resources \\
Input sequence length ($k$) & 100 \\
Output sequence length ($l$) & 10 \\
Output dimension & 1 \\
\hline
\multicolumn{2}{|c|}{\textbf{DQN-specific Parameters}} \\
\hline
Discount factor & 0.9 \\
Epsilon & Decaying strategy \\
Rewards & 1 (correct prediction); -1 (otherwise) \\
\hline
\multicolumn{2}{|c|}{\textbf{OLF Parameters}} \\
\hline
$\alpha$ & 10 \\
$q_{th}$ & 0.5 \\
\hline
\end{tabular}
\end{table}
\begin{figure*}[htbp]    \centering\includegraphics[trim= 2.5cm 0.1cm 0.9cm 2.3cm, clip, width=\textwidth]{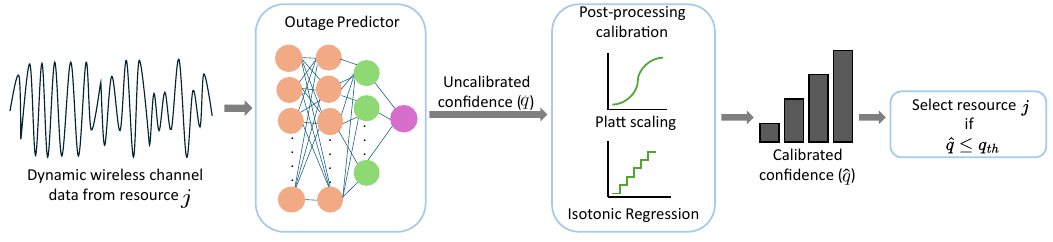}    \caption{System-level overview of the ML-based outage prediction framework. The predictor takes dynamic wireless channel data for resource $j\in \mathcal{R}$ as input and produces uncalibrated confidence estimates. Post-processing calibration yields calibrated confidence values, which are then used for threshold-based resource selection.}    \label{fig:system_model}\end{figure*}
\subsection{Data Generation}
The training/testing dataset for the channel is synthetically obtained as follows. We start by generating a zero-mean complex Gaussian vector given by: $G[0] = [g_1, g_2, \dots, g_v]$
at $t = 0$, with $v = 1024$, where each element has a variance of $\frac{1}{v}$. To simulate the motion of a mobile user, at each time step $t$, we apply a per-element, independent and uniformly distributed phase shift $\theta_{it} \sim \text{Unif}\left(-\frac{\pi}{2}, \frac{\pi}{2}\right)$.
At time $t = 1$, we have:
\begin{equation}
G[1] = \Bigg[e^{i\xi \sin\theta_{11}}g_1, e^{i\xi \sin\theta_{21}}g_2, \ldots, e^{i\xi \sin\theta_{v1}}g_v\Bigg],
\end{equation} 
and so on. In the frequency domain, we have:

\begin{align}    
\!\mathcal{H}[1] \!= \!\frac{1}{\!\sqrt{v}} \!\Bigg[ 
&\!\sum_{m=0}^{v-1} \!e^{i\xi \!\sin\theta_{\!m1}} \!g_m \!e^{\!-i2\!\pi \!m/v}, \!\ldots, \sum_{\!m=0}^{\!v-1} \!e^{i\xi \!\sin\theta_{\!m1}}\! g_m \!e^{\!-i2\!\pi \!m}
\!\Bigg].
    \end{align}
Let $h_i(t)$ denote the $i$-th element of $\mathcal{H}[t]$. The channel for resource $j$ at time $t$ is modeled as:
\begin{equation}
y_j[t] = h_i[t] x_j[t] + w_j[t],
\end{equation} 
where $x_j[t]$ represents a unit variance signal term, $w_j[t]$ represents a unit variance zero mean complex Gaussian noise term, and $h_i[t]$ are stationary and zero mean complex Gaussian variables. Furthermore, the autocorrelation between successive channel samples is given by:
\begin{align}    
\mathbb{E}[h_j(t) h_j(t+1)] &= \mathbb{E} \left[ \frac{1}{v} \sum_{m=0}^{v-1} e^{i \xi \sin\theta_{m1}} \lVert g_m \rVert^2 \right] \nonumber\\
&= \mathbb{E} \lVert g_m \rVert^2 \mathrm{J_{0}\left(\xi\right)},
\label{eq:char_fun}
\end{align}
where $\mathrm{J_{0}(\cdot)}$ is the zeroth-order Bessel function of the first kind, and~\eqref{eq:char_fun} follows from the characteristic function of the uniform distribution. This model is in line with the widely accepted Clarke's 2D model\footnote{Additionally, channel data was generated using Clarke’s 3D model~\cite{618205}, where the autocorrelation is given by $\mathrm{sinc}(\cdot)$, in contrast to Clarke's 2D model used in this paper, where the autocorrelation is given by $\mathrm{J_0}(\cdot)$. This was done to support the general validity of our theoretical results across different input data distributions. The findings discussed in the following subsections were found to be consistent and equally applicable to this dataset.}~\cite{6779222}, where $\xi$ represents the displacement of the mobile receiver, thereby capturing the temporal correlation introduced by user mobility. This model ensures that the simulated fading process reflects realistic channel dynamics, enabling a meaningful evaluation of the outage predictor’s calibration performance.

\subsection{Training and Evaluation}
We utilize an ML-based outage predictor and compare two model architectures: a Deep Q-Network (DQN) incorporating an LSTM layer, as implemented in~\cite{10469572}, and a standalone LSTM model, as implemented in~\cite{10443669}. Both predictors are trained on channel data generated using Clarke’s 2D model, employing the OLF from~\cite{10443669} as well as the BCE loss function. The predictors are developed using TensorFlow’s Keras API and trained with the ADAM optimizer. Table~\ref{table1} details the architecture and hyperparameters of both the ML models used in our experiments. 

We train our predictors on a dataset comprising 4,500 × (number of resources) samples. For a 4-resource system, this corresponds to 18,000 training samples, with the dataset size scaling proportionally (e.g., 27,000 samples for a 6-resource system). The training and validation datasets are identical in size, and testing is performed on 11,000 instances of the $|\mathcal{R}|$-resource system to achieve statistically robust results. The predictor is retrained 10 times, and the average performance is used as a single data point for analysis. Each experiment generates sequences of $k + l$ channel samples per resource, where $k = 100$, $l = 10$, and $\xi = 0.1$ radians. The first $k$ samples are provided as input to the ML model, while the last $l$ samples, $\accentset{\rightarrow}{H}_{j}^l{(t)}$, are used to determine whether communication can be supported at a specified rate.
On a desktop computer equipped with a 12th Gen Intel(R) Core(TM) i9-12900HK, 2500 MHz, 14 Core(s) and 20 Logical processor(s), the training of a single predictor takes approximately 5 minutes. 

While post-processing calibration helps improve the reliability of predicted confidence scores, it introduces a slight computational overhead since it is applied separately after model training. However, the NN used in this study is lightweight and enables fast training, so the additional cost of calibration remains minimal, especially when compared to large-scale models with extensive parameter counts. To reduce computational overhead in practical implementations, calibration objectives can be integrated directly into the training process~\cite{karandikar2021softcalibrationobjectivesneural,mukhoti2020calibratingdeepneuralnetworks}. By eliminating the need for separate post-processing calibration, this approach can be more suitable for deployment in energy-constrained or real-time systems.

The reliability diagrams presented in Section~\ref{sec:results} are constructed using logarithmic binning as defined in~\cite[Eq. (7)]{10757632}: \begin{equation} \left(0,1\right] = \bigcup_{n = 0}^{\infty}\left[ b^{-(n+1)}, b^{-n} \right], \label{eq:log_bin} \end{equation} where $b = \sqrt{10}$. This binning scheme enables the assessment of calibration behavior across several orders of magnitude in predicted confidence, with enhanced resolution in the low-confidence regions. This feature is particularly relevant in wireless communications, where systems often operate in the very low regions of performance metrics such as OP, bit/symbol error rates, etc. These regimes correspond to high-reliability operation, where fine-grained calibration assessment becomes crucial.
These reliability plots represent the average calibration performance of our outage predictor over 10 runs, with each run evaluating calibration on 10,000 samples.

To empirically demonstrate the key properties of the OP established under perfect calibration in Section~\ref{sec: on_the_impact_of_calibration}, we apply two widely used post-processing calibration techniques: Platt scaling and isotonic regression. Their effectiveness in improving calibration is evaluated using two standard metrics: Negative Log-Likelihood (NLL)~\cite{guo2017calibration} and Brier Score (BS)~\cite{murphy1977reliability,brier1950verification}. NLL, also known as cross-entropy loss~\cite{lecun2015deep}, measures the alignment of predicted probabilities with true labels~\cite{kull2017beta}. The BS captures the mean squared difference between predicted probabilities and true outcomes, with values $\in [0, 1]$. For both the metrics, lower values indicate better calibration.

We now briefly describe the calibration techniques implemented using widely available Python libraries, including \textit{scikit-learn} for logistic and isotonic regression and \textit{SciPy} for mathematical computations, as detailed below.
\begin{enumerate}
\item \textit{Platt Scaling:} This is a popular method for calibrating binary classifiers due to its simplicity and effectiveness. It fits a logistic regression model to the output logits (raw outputs). For each sample \(i\), the calibrated probability is given by:
\begin{equation}
    \hat{q}_i = \sigma(mz_i + n),
    \label{eq:cal_prob_platt}
\end{equation}
where \(z_i\) is the logit, \(m\) and \(n\) are parameters learned during calibration, and \(\sigma(\cdot)\) is the sigmoid function. Since the ML models used here outputs probabilities through a sigmoid activation, we first convert them to logits using the \texttt{logit} function:
\begin{equation}
    \texttt{logit}(p) = \ln\left(\frac{p}{1-p}\right),
\end{equation}
where \(p\) is the predicted probability. These logits serve as inputs to scikit-learn's \texttt{LogisticRegression}, which fits the parameters \(m\) and \(n\) by minimizing the NLL on a validation set using the \texttt{lbfgs} solver. The NN’s parameters
remain fixed, ensuring calibration only adjusts the output
probabilities.
\item \textit{Isotonic Regression:} This is a non-parametric method that learns a non-decreasing, piecewise-constant function \(f(\cdot)\) to map uncalibrated probabilities \(\hat{p}_i\) to calibrated values \(\hat{q}_i = f(\hat{p}_i)\), where \(f(\cdot)\) is optimized to minimize the squared loss:
\begin{equation}
    \sum_{i=1}^n \left(f(\hat{p}_i) - y_i \right)^2,
\end{equation}
where \(y_i\) denotes the true label. We implement this using scikit-learn's \texttt{IsotonicRegression}, which fits the function using the pool-adjacent-violators algorithm. This ensures monotonicity while minimizing the squared error on the validation set. Once fitted, the model adjusts new predictions by applying the learned isotonic function to improve alignment with observed outcomes.
\end{enumerate}
\begin{figure}[t]
    \centering\includegraphics[trim= 0.6cm 1.3cm 0.6cm 1.3cm, clip, width=0.43\textwidth]{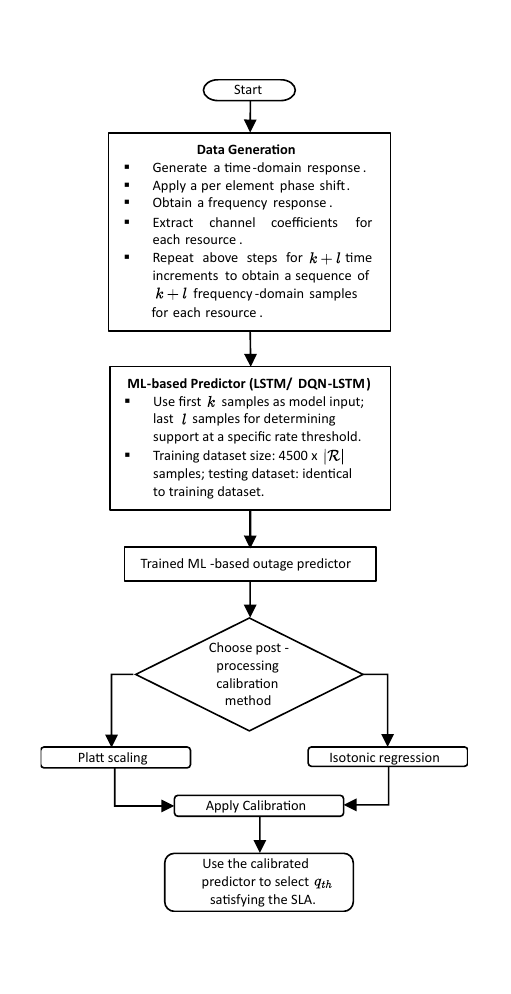}    \caption{Flowchart illustrating the key steps required to reproduce the proposed calibration-aware ML-based resource allocation framework.}
    \label{fig:flowchart}
\end{figure}
\figref{fig:system_model} shows a system-level overview of the ML-based outage prediction framework. The predictor processes dynamic wireless channel data for resource $j \in \mathcal{R}$ and outputs uncalibrated confidence estimates. These estimates are then calibrated through post-processing calibration methods, resulting in confidence values that are used to select resources based on a predefined classification threshold. Furthermore, a flowchart outlining the key steps involved in implementing the proposed ML-based resource allocation framework, including post-processing calibration, is shown in Fig.~\ref{fig:flowchart}.

Our study employs offline, post-processing calibration methods such as Platt scaling and isotonic regression. However, these static approaches may deteriorate under distribution shifts~\cite{ovadia2019trustmodelsuncertaintyevaluating}. These shifts can be mitigated using online adaptation, which updates NN parameters during deployment to adapt to evolving data distributions, or through robust strategies that improve calibration performance across diverse conditions~\cite{ouala2023onlinecalibrationdeeplearning}. Examples include architectural designs such as Annealing Double-Head networks~\cite{guo2023annealingdoubleheadarchitectureonline}, gradient-based updates via differentiable or approximated solvers, and gradient-free schemes like ensemble Kalman inversion~\cite{Iglesias2013}, all of which help sustain calibration under shifting conditions.
\section{Results and Demonstration}
\label{sec:results}
This section presents numerical results that demonstrate how the output of a calibrated predictor relates to the system’s OP and helps identify an appropriate $q_{th}$ to meet reliability targets. We begin by evaluating the predictor’s calibration using post-processing methods and reliability diagrams with logarithmic binning. This is followed by a robust simulation-based analysis that demonstrates the theoretical results established in Theorems~\ref{thm thm1} and~\ref{thm thm3}. Finally, we present a proof of concept for the resource allocation technique introduced in Section~\ref{sec:system_model}.


\begin{figure*}[!t]
    \centering
\includegraphics[trim= 0.07cm 13.1cm 0.1cm 0.6cm, clip, width=\textwidth]{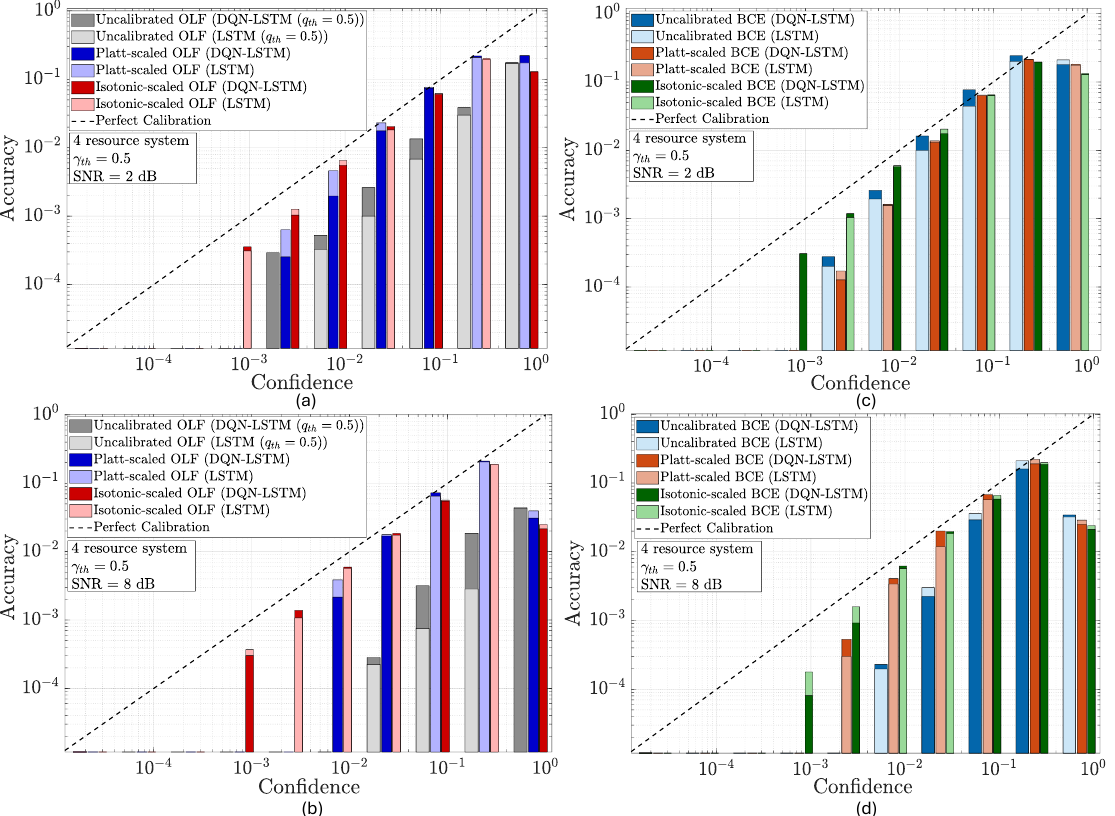}
    \caption{Histogram-based reliability diagrams of our predictor utilizing OLF and BCE for a 4-resource system with a rate threshold $\gamma_{th}$ of 0.5, using both LSTM and DQN-LSTM model architectures. 
    Specifically, plots are presented for $\mathtt{SNR}$ values of (a) 2~\rm{dB} and (b) 8~\rm{dB}, for predictors trained with OLF, and (c) 2~\rm{dB} and (d) 8~\rm{dB}, for predictors trained with BCE, each with various calibration methods applied.} 
    \label{fig:reliability_plots}
\end{figure*}
\subsection{Results}    
\figref{fig:reliability_plots} presents the histogram-based reliability plots using logarithmic binning for our outage predictor with a 4-resource system where $\gamma_{th} = 0.5$ and $q_{th} = 0.5$. The results are shown for both the LSTM and DQN-LSTM model architectures.
Fig.~\ref{fig:reliability_plots}~(a)~and~(b) show logarithmic reliability plots for predictors trained with OLF at an $\mathtt{SNR}$ of 2~\rm{dB} and 8~\rm{dB}, respectively, while Fig.~\ref{fig:reliability_plots}~(c) and~(d) present the corresponding logarithmic reliability plots for predictors trained with BCE. Furthermore, Table~\ref{tab:calibration_results_extended} presents the NLL and BS values for predictors trained with OLF and BCE, comparing the performance of calibrated and uncalibrated predictors across different $\mathtt{SNR}$ levels for both model architectures.

\begin{table*}[t]
    \centering
    \caption{NLL and BS Values across ML models and loss functions at different $\mathtt{SNR}$ Levels: Uncalibrated vs. Calibrated}
    \small
    \renewcommand{\arraystretch}{1.1}
    \setlength{\tabcolsep}{4pt} 
    \begin{tabular}{|c|c|c|c|c|c|c|c|c|c|c|c|c|}
        \hline
        \multirow{3}{*}{SNR} 
        & \multicolumn{6}{c|}{OLF (DQN-LSTM)} 
        & \multicolumn{6}{c|}{BCE (DQN-LSTM)} 
        \\ \cline{2-13}
        & \multicolumn{3}{c|}{NLL} & \multicolumn{3}{c|}{BS}
        & \multicolumn{3}{c|}{NLL} & \multicolumn{3}{c|}{BS}
        \\ \cline{2-13}
        & Uncalibrated & Platt & Isotonic & Uncalibrated & Platt & Isotonic 
        & Uncalibrated & Platt & Isotonic & Uncalibrated & Platt & Isotonic 
        \\ \hline
        2~\rm{dB}  & 0.33 & 0.12 & 0.08 & 0.09 & 0.05 & 0.02 & 0.09 & 0.09 & 0.09 & 0.03 & 0.03 & 0.03 \\ \hline
        8~\rm{dB}  & 0.18 & 0.05 & 0.03 & 0.06 & 0.03 & 0.01 & 0.05 & 0.04 & 0.04 & 0.01 & 0.01 & 0.01 \\ \hline
        \multirow{3}{*}{SNR} 
        & \multicolumn{6}{c|}{OLF (LSTM)} 
        & \multicolumn{6}{c|}{BCE (LSTM)} 
        \\ \cline{2-13}
        & \multicolumn{3}{c|}{NLL} & \multicolumn{3}{c|}{BS}
        & \multicolumn{3}{c|}{NLL} & \multicolumn{3}{c|}{BS}
        \\ \cline{2-13}
        & Uncalibrated & Platt & Isotonic & Uncalibrated & Platt & Isotonic 
        & Uncalibrated & Platt & Isotonic & Uncalibrated & Platt & Isotonic 
        \\ \hline
        2~\rm{dB}  & 0.36 & 0.10 & 0.08 & 0.09 & 0.03 & 0.02 & 0.08 & 0.08 & 0.07 & 0.03 & 0.03 & 0.03 \\ \hline
        8~\rm{dB}  & 0.22 & 0.05 & 0.02 & 0.05 & 0.01 & 0.01 & 0.04 & 0.03 & 0.03 & 0.03 & 0.03 & 0.03  \\ \hline
    \end{tabular}
    \label{tab:calibration_results_extended}
\end{table*}

From Fig.~\ref{fig:reliability_plots}, it is evident that the application of both the calibration methods significantly enhances the calibration performance of the outage predictor. This improvement is not only observed visually but also supported by the lower NLL and BS values achieved after calibration. Notably, both the LSTM and DQN-LSTM model architectures achieve similar performance, highlighting the effectiveness of the calibration methods across both models. To better understand their impact, we assess the effectiveness of each calibration method on the OLF-trained outage predictor below. 
\begin{enumerate}
    \item \textbf{Platt Scaling:} 
   This method improves calibration, as shown in~\figref{fig:reliability_plots} (a) for $\gamma_{th} = 0.5$ and $\mathtt{SNR} = 2~\mathrm{dB}$, where the dark grey bars represent the uncalibrated results and the dark blue bars represent the Platt scaled results, obtained using the DQN-LSTM model. The calibration improvement is also reflected in the reduced NLL and BS values after calibration (see Table~\ref{tab:calibration_results_extended}), with NLL decreasing from 0.33 (uncalibrated) to 0.12 (Platt scaled) and BS dropping from 0.09 (uncalibrated) to 0.05 (Platt scaled). In general, this method demonstrates its effectiveness primarily at higher confidence levels, particularly within the range of $10^{-2}$ to $1$.
    \item \textbf{Isotonic Regression:} This method provides significant calibration improvements through non-parametric adjustments to predicted probabilities. For example, in~\figref{fig:reliability_plots}~(b), for $\gamma_{th}=0.5$ and \(\mathtt{SNR} = 8~\mathrm{dB}\), the light grey bars represent the uncalibrated results and the light red bars represent the isotonic scaled results, using the LSTM model. Here, the light red bars align closely with the perfect calibration line, illustrating its effectiveness. This is further validated by a notable reduction in NLL from 0.22 (uncalibrated) to 0.02 (isotonic scaled) and in BS from 0.05 (uncalibrated) to 0.01 (isotonic scaled). It can be observed that this method proves to be effective even at lower confidence levels. 
\end{enumerate}
\setlength{\abovecaptionskip}{-1.3pt}
\begin{figure}[!t]
    \centering
\includegraphics[trim= 0.3cm 14.5cm 2.0cm 0.6cm, clip, width=\textwidth]{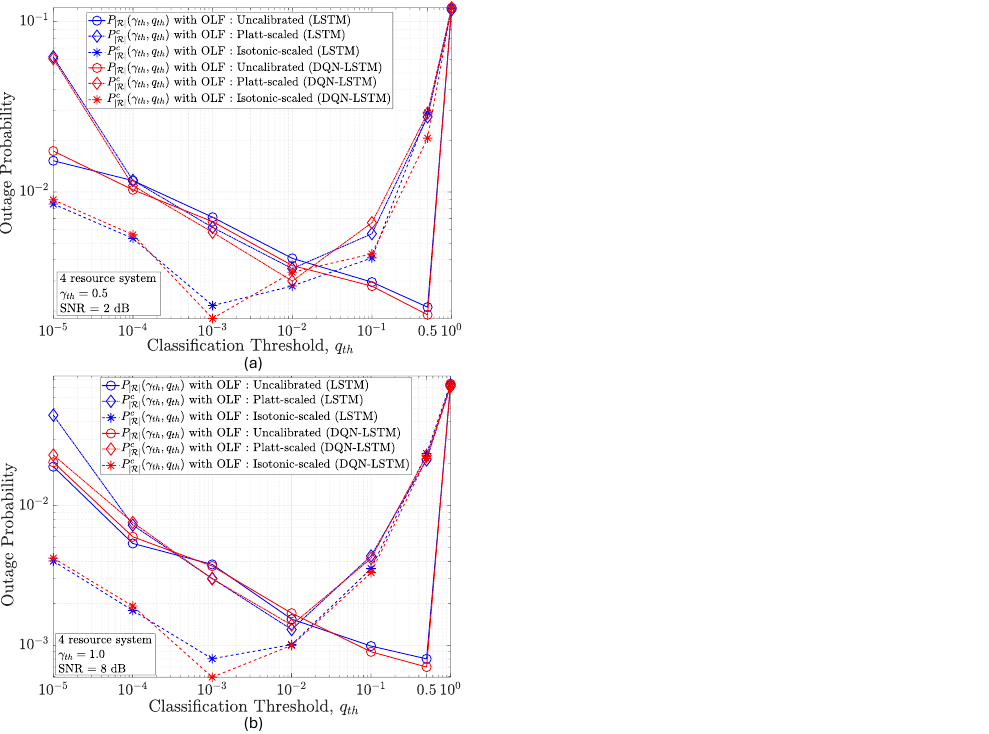}    \caption{$P^\star_{\vert\mathcal{R}\vert}(\gamma_{th}, q_{th})$ vs $q_{th}$ for a 4-resource system using OLF, evaluated for both LSTM and DQN-LSTM architectures, with (a) $\gamma_{th}~= 0.5$, at $\mathtt{SNR}$ of 2~\rm{dB} and (b) $\gamma_{th} = 1.0$, at $\mathtt{SNR}$ of 8~\rm{dB}.}
    \label{fig:OP_qth_plots_OLF}
\end{figure}

\setlength{\abovecaptionskip}{-1.3pt}
\begin{figure}[!t]
    \centering
\includegraphics[trim= 0.3cm 14.25cm 2.0cm 0.83cm, clip, width=\textwidth]{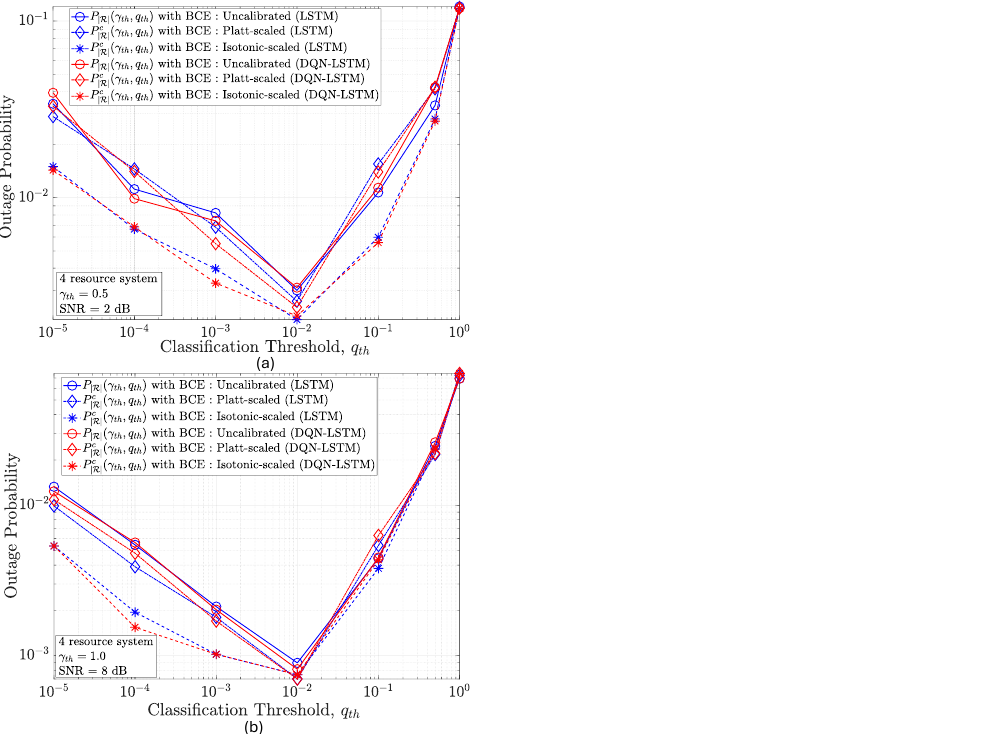}    \caption{$P^\star_{\vert\mathcal{R}\vert}(\gamma_{th}, q_{th})$ vs $q_{th}$ for a 4-resource system using BCE, evaluated for both LSTM and DQN-LSTM architectures, with (a) $\gamma_{th}~= 0.5$, at $\mathtt{SNR}$ of 2~\rm{dB} and (b) $\gamma_{th} = 1.0$, at $\mathtt{SNR}$ of 8~\rm{dB}.}
    \label{fig:OP_qth_BCE}
\end{figure}
Platt scaling, being parametric, is efficient and less prone to overfitting when the held-out validation data is limited, whereas isotonic regression, as a non-parametric method, offers greater flexibility and achieves better calibration with sufficient data. In our setting, where accuracy across a broader confidence range is critical, isotonic regression consistently delivers superior results, as reflected by its lower NLL and BS values.

From~\figref{fig:reliability_plots} (c), and (d), it can be seen that the predictors trained with BCE display better calibration in their uncalibrated state compared to those trained with OLF. This is because BCE-based training is independent of $q_{th}$, enabling it to model the overall channel distribution. Post calibration, the predictors trained with BCE showed minimal changes, as reflected by the NLL and BS values (see Table~\ref{tab:calibration_results_extended}).

Figs.~\ref{fig:OP_qth_plots_OLF} and~\ref{fig:OP_qth_BCE} presents the OP for a 4-resource system using OLF and BCE, respectively, across a range of $q_{th}$ values, for (a) $\gamma_{th} = 0.5$ at $\mathtt{SNR} = 2~\mathrm{dB}$ and (b) $\gamma_{th} = 1.0$ at $\mathtt{SNR}~=~8~\mathrm{dB}$, using both LSTM and DQN-LSTM model architectures. Unlike Fig. 5 from~\cite{10443669}, these plots use $q_{th}$ values on a logarithmic scale and also present the OP obtained using the calibrated predictors. It can be seen that the OP obtained using OLF and BCE exhibit a ‘U'-shaped curve, indicating that both very small and very large $q_{th}$ values are not good. This is because very small $q_{th}$ values result in the user almost always being allocated the final resource, while very large $q_{th}$ values lead to allocation to the first resource. 
Notably, at $\!q_{th} \!= \!0.5$, the uncalibrated OLF-trained predictor outperforms the BCE-trained ones. As $q_{th}$ decreases, this performance gap narrows, and in some cases, BCE slightly outperforms OLF at lower $q_{th}$.

It can also be seen from~\figref{fig:OP_qth_plots_OLF} that the calibrated and uncalibrated OLF-trained predictors both achieve the same minimum OP but at different $q_{th}$ values.
For instance, in a 4-resource system with $\gamma_{th} = 0.5$ and $\mathtt{SNR} = 2~\mathrm{dB}$ (see~\figref{fig:OP_qth_plots_OLF}~(a)), the uncalibrated predictor trained with OLF achieves its minimum OP at $q_{th} = 0.5$, which matches its training threshold, while the calibrated predictor reaches the same minimum OP at $q_{th} = 10^{-3}$. 
This observation aligns with the result of Theorem~\ref{thm thm3}, which establishes that post-processing calibration cannot reduce the minimum achievable OP. It is important to note that this observation is not reflected in Fig.~\ref{fig:OP_qth_BCE}, because BCE-trained predictors are inherently well calibrated, as evidenced by~\figref{fig:reliability_plots} (c) and (d), and Table~\ref{tab:calibration_results_extended}. As previously mentioned, the distinction between the two training approaches highlights key differences in their calibration behavior. BCE-trained predictors generally exhibit better inherent calibration, as their training is independent of 
$q_{th}$ and models the entire channel distribution, whereas OLF is specifically designed to optimize outage performance at a fixed classification threshold.


\setlength{\abovecaptionskip}{-3pt} 
\begin{figure}[!t]
    \centering
\includegraphics[trim= 0.11cm 20.4cm 0.3cm 0.19cm, clip, width=\textwidth]{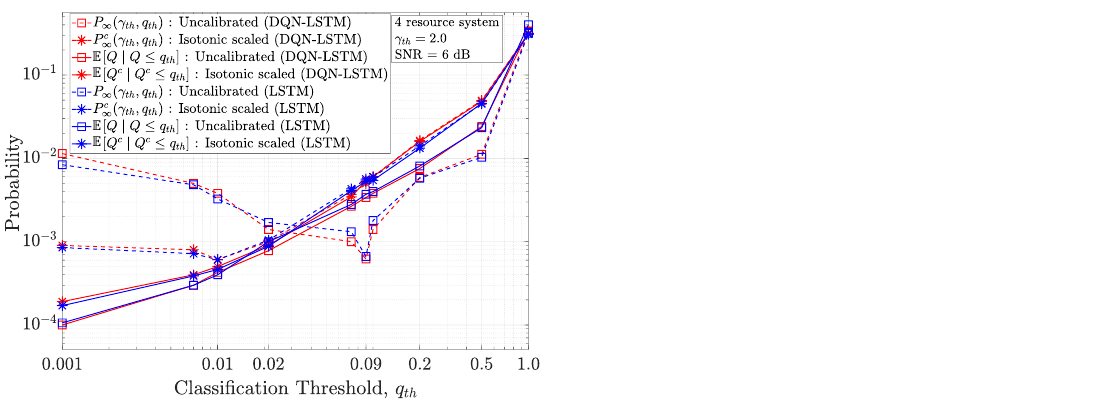}
    \caption{Comparison of $\!P^\star_{\infty}(\gamma_{th}, q_{th})$ and $\mathbb{E}[Q^\star\! \mid Q^\star \leq q_{th}]$ across different $q_{th}$ values, in a 4-resource system using OLF, evaluated using both LSTM and DQN-LSTM architectures, with $\gamma_{th} \!= 2.0$ and $\mathtt{SNR} = 6\mathrm{dB}$.}
    \label{fig:Pinf_eqth}
\end{figure}

\begin{figure}[!t]
    \centering
\includegraphics[trim= 0.25cm 20.48cm 0.5cm 0.17cm, clip, width=\textwidth]{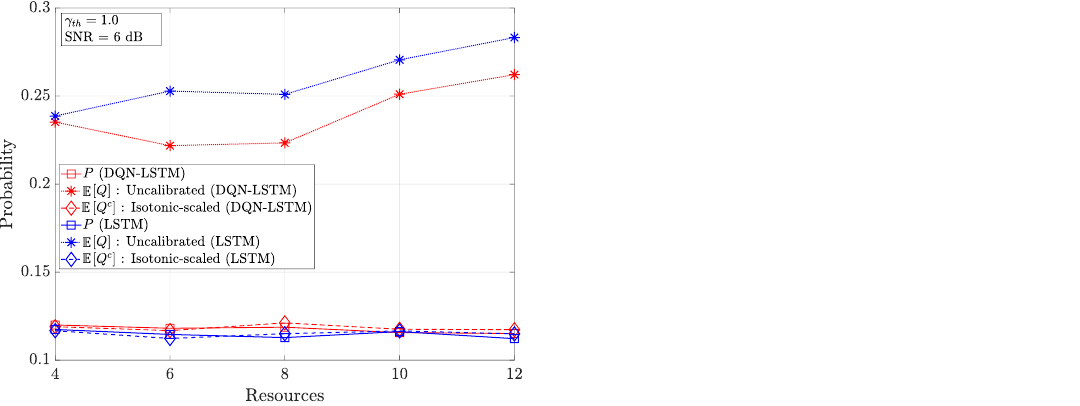}
    \caption{Comparison of $P$ and $\mathbb{E}[Q^\star]$ across different numbers of resources $\mathcal{R}$, using OLF, evaluated using both LSTM and DQN-LSTM architectures, with $\gamma_{th} = 1.0$ and $\mathtt{SNR} = 6~\mathrm{dB}$.}
    \label{fig:P1_EQ}
\end{figure}

Figs.~\ref{fig:Pinf_eqth} and~\ref{fig:P1_EQ} illustrate the key results\footnote{These plots do not compare the outage performance of OLF with BCE, as BCE-trained predictors are already well calibrated in their uncalibrated state (see Fig.~\ref{fig:reliability_plots} (c) and (d), and Table~\ref{tab:calibration_results_extended}).} discussed in Section~\ref{sec: on_the_impact_of_calibration}, demonstrating the behaviour established for a perfectly calibrated predictor in Theorem~\ref{thm thm1}, using both LSTM and DQN-LSTM models. For this analysis, we apply isotonic regression, as it consistently achieves better calibration performance, according to the NLL and BS metrics, compared to Platt scaling. In particular, Fig.~\ref{fig:Pinf_eqth} shows the OP for an infinite resource limit, $P^\star_{\infty}(\gamma_{th}, q_{th})$ alongside the expected confidence of the predictor when its output is below the threshold $q_{th}$, i.e., $\mathbb{E}[Q^\star \!\mid \!Q^\star \leq q_{th}]$. The figure compares calibrated and uncalibrated predictors over a range of $q_{th}$ values, using a system with 4 resources, $\gamma_{th} = 2.0$, and $\mathtt{SNR} = 6~\mathrm{dB}$. It can be observed that these curves align closely for the calibrated predictor when isotonic regression is applied, consistent with the behavior described in Theorem~\ref{thm thm1}, Eq.~\eqref{eq:Pinf_equal_Eqth_thm1}. 

Furthermore, this result helps in selecting an appropriate $q_{th}$ to meet a target system OP. For instance, a service provider might have a service level agreement (SLA) requiring the OP to be no greater than $0.01$ (i.e., a user should experience an outage no more than once every 100 transmissions). To satisfy this requirement, the provider can consult the calibrated outage predictor\footnote{
This example applies only to a calibrated predictor, for which~\eqref{eq:Pinf_equal_Eqth_thm1} holds. In the uncalibrated case, the output confidence does not reflect the true OP; this equation no longer holds, and $q_{th}$ must be chosen via empirical search.}.
Given $\gamma_{th} = 2.0$, $\mathtt{SNR} = 6~\mathrm{dB}$, and a sufficient number of resources, Fig.~\ref{fig:Pinf_eqth} shows that setting $q_{th} = 0.2$ for a well-calibrated predictor satisfies this SLA. 
Additionally, we observe that both the calibrated and uncalibrated OLF-trained predictors achieve the same minimum OP, but at different $q_{th}$ values. The shift in the $q_{th}$ value where the calibrated predictor achieves its minimum OP is a consequence of calibration, which adjusts the predicted confidence scores to better reflect true OPs thereby rescaling the threshold at which the minimum OP occurs.

\figref{fig:P1_EQ} compares the single-resource OP, $P$, with the expected predictor output, $\mathbb{E}[Q^\star]$, evaluated at~$\gamma_{th} = 1.0$ and $~\mathtt{SNR}~= 6~\mathrm{dB}$ for different number of resources. It can be seen that the calibrated predictor produces $\mathbb{E}[Q^c]$ values that closely match the observed single-resource OP $P$, consistent with Theorem~\ref{thm thm1}, Eq.~\eqref{eq:P_equals_EQ}.
\subsection{Proof of Concept}
\begin{figure*}[!t]
    \centering
\includegraphics[trim= 1.3cm 15.0cm 2.0cm 0.83cm, clip, width=\textwidth]{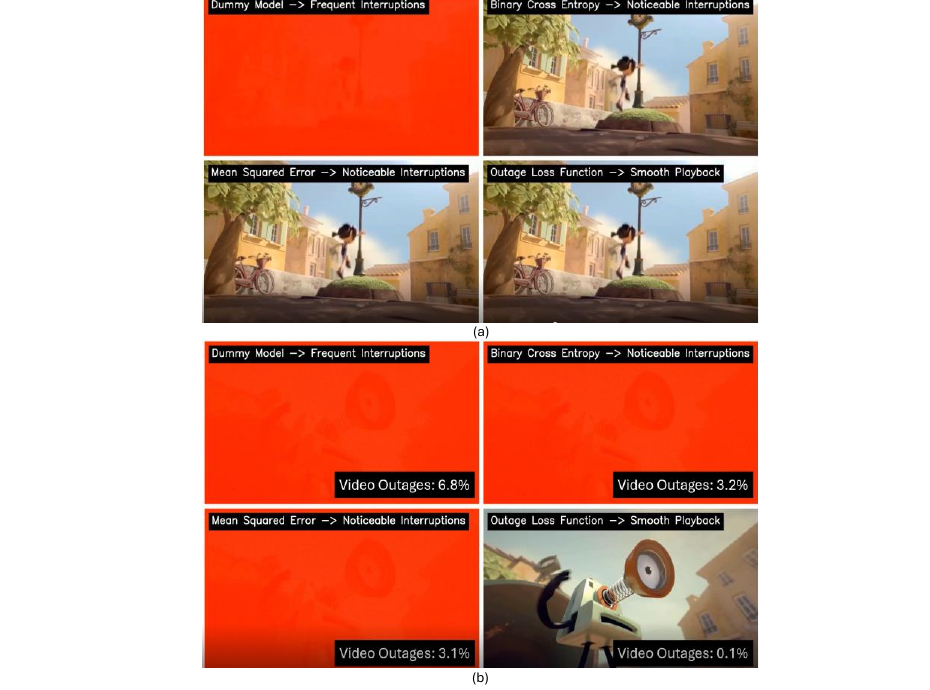}    \caption{Snapshots from the video demonstration comparing a dummy model (non-ML strategy; top left) with ML-based outage predictors trained using BCE (top right), MSE (bottom left), and the OLF (bottom right). (a) Red frames indicate instances where the predictor incorrectly recommended a channel to support the video, while smooth playback reflects more effective channel selection. (b) The average outage rate over the entire video duration corresponding to each predictor.}
    \label{fig:demo_video_1}
\end{figure*}

To demonstrate the application of the resource allocation technique and the OLF discussed in Section~\ref{sec:system_model},~\cite{github} showcases video transmission over fading wireless channels using four different outage predictors. The goal is to compare the effectiveness of these predictors in reducing the likelihood of video outages and ensuring smooth playback. In this setup, a resource refers to a wireless channel used for video transmission. When a channel’s signal quality degrades due to fading and can no longer support reliable video transmission, an outage is likely to occur. However, the system leverages the outage predictor to anticipate such events and proactively switch to a more reliable channel. Playback is interrupted only when a predictor incorrectly recommends a channel that fails to meet the transmission requirement, in which case the screen turns red to indicate the outage (see Fig.~\ref{fig:demo_video_1} (a)).


For this demonstration, we set $\gamma_{th} = 2.2$, and $\mathtt{SNR}$ = 15~$\mathrm{dB}$. The video runs at 25 frames per second, with predictions made 80 ms ahead, which is equivalent to two frames. The predictor is trained to identify whether an outage will occur within the next two frames, using either the BCE, MSE, or OLF. A baseline, non-ML strategy, referred to as a dummy model, is also included. This model always selects the first channel regardless of outage conditions.


This demonstration reinforces the effectiveness of the OLF-trained predictor, which resulted in an average outage rate of only 0.1\% over the entire video duration, significantly lower than that of the dummy model (6.8\%), and the predictors trained using BCE (3.2\%) and MSE (3.1\%), leading to smoother video playback. Please note that the demonstration here uses the uncalibrated predictor with $q_{th}$ set to 0.5. However, it is worth highlighting that the calibrated predictor, trained using the OLF, will also achieve at best the same OP performance through the selection of an appropriate $q_{th}$, in accordance with Theorem~2 presented in this work. This behavior is consistent with Fig.~\ref{fig:OP_qth_plots_OLF}, which shows that both the calibrated and uncalibrated predictors trained with OLF achieve the same minimum system OP, but at different $q_{th}$ (classification threshold) values.


\section{Conclusion}
\label{sec:conclusion}
This paper studies the calibration performance of the outage predictor used in an ML-assisted resource allocation system. A preliminary proof of concept demonstrating the application of the resource allocation technique referred to here, aimed at a broader audience, can be found in~\cite{github}. In this paper, we established key theoretical properties of the OP under perfect calibration. Specifically, we demonstrated that, in the infinite-resource case, the OP corresponded to the expected predictor output conditioned on it being below the classification threshold, while in the single-resource case, it equaled the predictor’s overall expected confidence. We then presented OP conditions for a perfectly calibrated predictor. These findings help guide the service provider in selecting $q_{th}$ values that satisfy the reliability requirements specified in an SLA. For instance, if an SLA permits no more than 1 outage in 200 transmissions (i.e., a system OP of $5 \times 10^{-3}$), system engineers can rely on a well-calibrated predictor to meet this requirement by setting $q_{th} = 0.09$ (see Fig.~\ref{fig:Pinf_eqth}). 

Notably, we showed that post-processing calibration cannot improve the system’s minimum achievable OP. Although both calibrated and uncalibrated predictors can achieve the same minimum OP, only calibrated predictors produce confidence scores that accurately reflect the true outage likelihood, enabling appropriate selection of $q_{th}$ values. In contrast, for uncalibrated predictors, $q_{th}$ must be tuned empirically. Furthermore, we identified a monotonicity-based condition on the predictor’s accuracy-confidence function that ensures improvement in the system’s OP, while also identifying a critical point where it remains unchanged.

Several fronts of new investigations can be envisaged for future work. One of them shall focus on exploring conformal prediction as an alternative to post-processing calibration methods. Unlike methods that recalibrate predicted confidence scores to better align with observed outcome frequencies, conformal prediction provides formal calibration guarantees, ensuring that the true outcome falls within a set of possible outcomes with a user-specified confidence level. This makes it particularly appealing in scenarios involving distribution shift. Another one shall explore systems operating in fading scenarios other than the Rayleigh one to understand how worse or better propagation conditions impact the calibration performance. A preliminary investigation appears in~\cite{raina2025ml}. On another front, the aim is to extend the single-user approach used here to a multiuser-multirate approach, which better conforms to more practical and advanced networks.
\appendices
\appendix
\section{Appendix}	
\subsection{Proof of Theorem~\ref{thm thm1}  \label{app:thm1_proof}}
We consider the proof of this theorem in three parts.

Before proceeding, we recall that the accuracy-confidence function associated with the predictor is defined as:
\begin{equation}
A(q) \triangleq \mathbb{P} \left[ C< \gamma_{th} \mid Q^\star = q \right],
\label{eq:accuracy_confidence_recall}
\end{equation}
where perfect calibration corresponds to
\begin{equation}
A(Q^c) = Q^c.
\label{eq:AQ_Q_equals_Q}
\end{equation}
\begin{enumerate}
\item \textbf{For a single-resource system: }
Recall from~\eqref{eq:outage_probability_for_singe_resource}, the OP of a single resource system is given by:
\begin{equation}
P \triangleq \mathbb{P}  \left[C < \gamma_{th}\right].\nonumber
\end{equation}
Using the law of total expectation:
\begin{align}
    P &= \mathbb{E}_{Q^\star}\left[\mathbb{P}\left[C < \gamma_{th} \mid Q^\star\right]\right]
    \label{eq:law_of_total_expectation}\\
    &= \mathbb{E}_{Q^\star}[A(Q^\star)].
    \label{eq:proof_thm1_1}
\end{align}
For perfect calibration, we obtain
\begin{equation}
    P = \mathbb{E}[Q^c]
    \label{eq:P_equals_EQ_final_proof},
\end{equation}
where~\eqref{eq:P_equals_EQ_final_proof} follows from~\eqref{eq:AQ_Q_equals_Q}.

\item \textbf{For ${\vert \mathcal{R} \vert} \rightarrow \infty$}:
    The OP in the infinite resource limit is given by:
\begin{align}
P^\star_{\infty}(\gamma_{th}, q_{th}) &= \mathbb{P}\left[C < \gamma_{th} \mid Q^\star \leq q_{th}\right] \label{eq:OP_infty_general} \\
&= \mathbb{E}_{Q^\star \leq q_{th}} \left[ \mathbb{P} \left[ C < \gamma_{th} \mid Q^\star \right] \right] \label{eq:E_P_C_Q} 
\end{align}
\begin{align}
&= \mathbb{E}_{Q^\star \leq q_{th}} [A(Q^\star)]. \label{eq:OP_infty_accuracy_function}
\end{align}
For perfect calibration, we obtain
\begin{equation}
  P^c_{\infty}(\gamma_{th}, q_{th}) = \mathbb{E}[Q^c \mid Q^c \le q_{th}]. \label{eq:Pinf_equal_Eqth}
\end{equation}
Here,~\eqref{eq:E_P_C_Q} is given by the expectation taken over the predictor's output $Q^\star \leq q_{th}$. Equation~\eqref{eq:OP_infty_accuracy_function} follows from~\eqref{eq:accuracy_confidence_recall} and~\eqref{eq:Pinf_equal_Eqth} follows from~\eqref{eq:AQ_Q_equals_Q}.

\item \textbf{For an $|\mathcal{R}|$ resource system:} Substituting~\eqref{eq:Pinf_equal_Eqth} and~\eqref{eq:P_equals_EQ_final_proof} in ~\cite[eq. (13)]{10443669}, we obtain~\eqref{eq:PR_expression}. This completes the proof.
\end{enumerate}

\subsection{Proof of Theorem~\ref{thm thm3}  \label{app:thm3_proof}}
Recall from~\eqref{eq:Qc = rQ} we have $Q^c=\;
r\circ Q,$
\begin{align}
\implies& P^c_{|\mathcal{R}|}(\gamma_{th}, q_{th}) = \mathbb{P}\left(C < \gamma_{th} \mid Q^c \leq q_{th}\right), \label{eq:OP_def} 
\\
\implies&\min_{q_{th}} P^c_{|\mathcal{R}|}(\gamma_{th}, q_{th}) \ge \min_{q_{th}} P_{|\mathcal{R}|}(\gamma_{th}, q_{th}), \label{eq:OP_min_DPI}
\end{align}
where~\eqref{eq:OP_min_DPI} follows from the fact that, minimizing OP over \( Q^c \) corresponds to minimizing over a subset of the thresholds available through \( Q \). This completes the proof.

\subsection{Proof of Theorem~\ref{thm thm2}  \label{app:thm2_proof}}

We first show that $\mathbb{E}[A(Q^\star) \mid Q^\star\leq q_{th}] < P$ implies
$\mathbb{E}[A(Q^\star) \mid Q^\star \leq q_{th}] < \mathbb{E}[A(Q^\star) \mid Q^\star > q_{th}]$:
\begin{align}
   \noindent \quad 
    \mathbb{E}[A(Q^\star) \mid Q^\star \leq q_{th}] &< P \nonumber \\
    &= \mathbb{E}[A(Q^\star) \mid Q^\star \leq q_{th}] F(q_{th})\nonumber\\ &+\mathbb{E}[A(Q^\star) \mid Q^\star > q_{th}] (1 - F(q_{th})) 
    \label{eq:thm2_proof_eq1}
\end{align}
\begin{align}
     \implies\
    &\mathbb{E}[A(Q^\star) \mid Q^\star \leq q_{th}] - \mathbb{E}[A(Q^\star) \mid Q^\star \leq q_{th}] F(q_{th}) \nonumber\\&~~~~~~~~~~~~< \mathbb{E}[A(Q^\star) \mid Q^\star > q_{th}] (1 - F(q_{th})) \label{eq:thm2_proof_eq2}
    \\
    \implies\ 
    &\mathbb{E}[A(Q^\star) \mid Q^\star \leq q_{th}] (1 - F(q_{th})) \nonumber \\
    &~~~~~~~~~~~~< \mathbb{E}[A(Q^\star) \mid Q^\star > q_{th}] (1 - F(q_{th})) 
    \end{align}
    \begin{align}
    \implies\ 
    &\mathbb{E}[A(Q^\star) \mid Q^\star \leq q_{th}] 
    < \mathbb{E}[A(Q^\star) \mid Q^\star > q_{th}]
\end{align}
 where~\eqref{eq:thm2_proof_eq1} follows from the law of total expectation, and~\eqref{eq:thm2_proof_eq2} follows from rearrangement.
 
\noindent Next, we show that $\mathbb{E}[A(Q^\star)\! \mid Q^\star \!\leq q_{th}]\! < \mathbb{E}[A(Q^\star) \mid Q^\star > \!q_{th}]$ implies $\mathbb{E}[A(Q^\star) \mid Q^\star \leq q_{th}] < P$:
 \begin{align}
     P &= \mathbb{E}[A(Q^\star)]\nonumber\\
     &=\mathbb{E}[A(Q^\star) \mid Q^\star \leq q_{th}] F(q_{th})\nonumber\\ &~~~+\mathbb{E}[A(Q^\star) \mid Q^\star > q_{th}] (1 - F(q_{th})) \\
     &>\mathbb{E}[A(Q^\star) \mid Q^\star \leq q_{th}] F(q_{th})\nonumber\\&~~~+\mathbb{E}[A(Q^\star) \mid Q^\star \leq q_{th}] (1 - F(q_{th}))
     \label{eq:thm2_proof_eq3}\\
     &=\mathbb{E}[A(Q^\star) \mid Q^\star \leq q_{th}].
     \label{eq:thm2_proof_eq4}
 \end{align}
Finally, we consider the critical point, defined by\begin{equation}    \mathbb{E}[A(Q^\star) \mid Q^\star \leq q_{th}] = \mathbb{E}[A(Q^\star) \mid Q^\star > q_{th}].\end{equation}Combining this with the law of total expectation, we have\begin{equation}    \mathbb{E}[A(Q^\star) \mid Q^\star \leq q_{th}] = \mathbb{E}[A(Q^\star)]. \end{equation} Substituting this into~\eqref{eq:OP_case1_greedy_scheme} gives \begin{align} P^\star_{|\mathcal{R}|}(\gamma_{th}, q_{th}) &= \mathbb{E}[A(Q^\star)] (1 - F(q_{th}))^{|\mathcal{R}| - 1} \nonumber\\ &\quad ~~~+ \mathbb{E}[A(Q^\star)] \left(1 - (1 - F(q_{th}))^{|\mathcal{R}| - 1} \right) \nonumber\\ &= \mathbb{E}[A(Q^\star)], \end{align} meaning that the predictor yields no gain in the outage performance. This completes the proof.

\bibliographystyle{IEEEtran}
\bibliography{IEEEabrv,ref}

\end{document}